\documentclass[letterpaper, 10 pt, journal]{ieeeconf}  

\IEEEoverridecommandlockouts                              

\usepackage{mfirstuc}
\usepackage{graphics} 
\usepackage{epsfig} 
\usepackage{mathptmx} 
\usepackage{times} 
\usepackage{amsmath} 
\usepackage{amssymb}  
\usepackage{subcaption}
\usepackage[font={small}]{caption}
\usepackage{siunitx}
\usepackage{verbatim}
\usepackage[noadjust]{cite}
\usepackage{siunitx}
\usepackage{booktabs}
\usepackage{url}
\usepackage{listings}
\usepackage[dvipsnames]{xcolor}
\usepackage[normalem]{ulem}
 \usepackage{ulem} 
 
\renewcommand{\baselinestretch}{1}
\title{\LARGE \bf Tactile Perception in Upper Limb Prostheses:\\Mechanical Characterization, Human Experiments,\\and  Computational Findings }

\author{Alessia S. Ivani\authorrefmark{5}$^{1,2,3}$, Manuel G. Catalano$^{1}$, Giorgio Grioli$^{1,2}$, Matteo Bianchi $^{2,3}$,\\Yon Visell $^{4}$, and Antonio Bicchi$^{1,2,3}$  
\thanks{*  This article has been accepted for publication in IEEE Transactions on Haptics. This is the author's version which has not been fully edited and
content may change prior to final publication. Citation information: DOI 10.1109/TOH.2024.3436827.
© 2024 IEEE.  Personal use of this material is permitted.  Permission from IEEE must be obtained for all other uses, in any current or future media, including reprinting/republishing this material for advertising or promotional purposes, creating new collective works, for resale or redistribution to servers or lists, or reuse of any copyrighted component of this work in other works.
This work was supported by the European Research Council Synergy Grant Natural BionicS (NBS) project (Grant Agreement No. 810346) and by
the Italian Ministry of Education and Research (MIUR) in the framework of
the FoReLab project and Crosslab project (Departments of Excellence); by
PNRR, M4 C2 I1.5 Ecosistema dell’Innovazione ”Tuscany Health Ecosystem
(THE)” - Ecosistema dell’innovazione sulle scienze e le tecnologie della
vita in Toscana (CUP I53C22000780001) - Spoke 9. }
\thanks{$^{1}$with  Soft Robotics for Human Cooperation and Rehabilitation, Istituto Italiano di Tecnologia, Genova 16163, Italy.}{}
\thanks{$^{2}$with Centro di ricerca E. Piaggio, University of Pisa, Pisa 56122, Italy.}
\thanks{$^{3}$with Department of Information Engineering, University of Pisa, Pisa 56122, Italy.}
\thanks{$^{4}$with Department of Mechanical Engineering and Bioengineering Program, University of California, Santa Barbara, USA.
}
 \thanks{\authorrefmark{5} Corresponding author \tt\small alessia.ivani@iit.it}
}
\begin{document}
\maketitle
\thispagestyle{empty}
\pagestyle{empty}

\begin{abstract}
Tactile feedback is essential for upper-limb prostheses functionality and embodiment, yet its practical implementation presents challenges. Users must adapt to non-physiological signals, increasing cognitive load. 
However, some prosthetic devices transmit tactile information through socket vibrations, even to untrained individuals. 
Our experiments validated this observation, demonstrating a user's surprising ability to identify contacted fingers with a purely passive, cosmetic hand. 
Further experiments with advanced soft articulated hands revealed decreased performance in tactile information relayed by socket vibrations as hand complexity increased.
To understand the underlying mechanisms, we conducted numerical and mechanical vibration tests on four prostheses of varying complexity.
Additionally, a machine-learning classifier identified the contacted finger based on measured socket signals.
Quantitative results confirmed that rigid hands facilitated contact discrimination, achieving 83\% accuracy in distinguishing index finger contacts from others.
While human discrimination decreased with advanced hands, machine learning surpassed human performance.  These findings suggest that rigid prostheses provide natural vibration transmission, potentially reducing the need for tactile feedback devices, which advanced hands may require.
Nonetheless, the possibility of machine learning algorithms outperforming human discrimination indicates potential to enhance socket vibrations through active sensing and actuation, bridging the gap in vibration-transmitted tactile discrimination between rigid and advanced hands.
\end{abstract}
\vspace{-0.5cm}
\section{INTRODUCTION}
The loss or absence of a hand deprives a person of multiple functions that drive interaction with the outside world.
Those obliviously include motor functions, communications, socialization and, prominently, sensory functions.
Most commercial prostheses do not provide a substitute for sensing despite recent research efforts and technological improvements in haptics.
Bensmaia et al. \cite{bensmaia2020restoration} present a comprehensive review of the current invasive and non-invasive methods and challenges encountered in developing sensory feedback systems for bionic hands.
There is experimental evidence that non-invasive upper limb sensory feedback prostheses would improve embodiment~\cite{FeedbackEverydayuse, marasco2011robotic} and reduce phantom limb pain~\cite{FeedbackEverydayuse,dietrich2012sensory}.
Nevertheless, the discrepancy between the several feedback solutions proposed in the literature~\cite{bensmaia2020restoration,nemah2019review} and the few commercially available sensorized bionic hands~\cite{akhtar2020touch} is evident because very few provide a substitution for cutaneous feedback~\cite{bensmaia2020restoration,akhtar2020touch}. 
One of the main technological difficulties behind that limit is the integration of sensors and actuators, which can compromise wearability and usability~\cite{review2014}.
Another fundamental obstacle is understanding how to provide relevant feedback that does not require all of a user's attention. 
Indeed, results about the effect of cutaneous feedback on prosthesis control performance are contradictory. Some highlight improvements, whereas others find no differences~\cite{sensinger2020review,FeedbackEverydayuse}. 
The challenge consists of balancing the quantity of information to be communicated.
Indeed, the execution of everyday tasks should provide meaningful information without confusing or increasing the conscious attention effort required to interpret signals \cite{Farina}.  
It should also be noted that the solutions proposed in the literature are evaluated in research laboratories rather than in daily living tasks, where the user has to manage significantly more exteroceptive and proprioceptive information~\cite{review2014}.

Limited research has explored the natural transmission of tactile signals through prosthetic devices despite its potential to offer insights into users' baseline perceptions and ultimately contribute to the refinement and optimization of haptic feedback systems.
During tool-extended sensing, Miller et al. \cite{miller2018sensing} proved that a rod mechanically transduces impact location into vibratory patterns decoded by the somatosensory system.
The mechanoreceptors in the human hand transduce the vibratory cues into neural response patterns that preserve the location-specifying information~\cite{miller2018sensing}.
Thus, the authors provide evidence of how handheld tools function as sensory extensions of the human body.
Likewise, as pointed out by Farina et al.~\cite{Farina}, a prosthetic user has natural sensory feedback in addition to vision and, depending on the type of prosthetic device, can 
hear and feel motor actuation, and perceive vibrations conveyed through the prosthetic socket. 
An evaluation of the threshold level for contact cues detection of socket-prosthetic limbs with respect to bone-anchored prostheses is performed in \cite{jacobs2000evaluation}.
The authors demonstrate that bone-anchored and socket-prosthetic limbs can activate stump-level receptors upon vibratory stimulation of the prosthetic thumb.
Furthermore, according to the recent study by Amoruso et al. \cite{amoruso2022intrinsic}, the existence of intrinsic somatosensory feedback by artificial body parts has been demonstrated to be crucial in enabling precise motor commands.
Thus, for prosthetic users, vibration patterns are important tactile information usually felt on the stump and partake in users' strategies to compensate for the lack of sensory information \cite{review2014,lundborg2001sensory}.

We posit that investigating natural tactile information transmission is crucial for developing intuitive, non-invasive haptic prosthetic devices, aiming to optimize conveyed information and reduce users' cognitive load.
In our experience, meaningful texture information is transmitted through socket-based prosthetic devices \cite{6755554,ivani2023vibes}.
During empirical observations, users exhibited the capability to discern minor tactile interactions on the digits of their prosthetic limb and, notably, to reconstruct the specific finger that had been contacted with a degree of reliability significantly exceeding random chance.
Considering the complexity of bionic hands, it is evident that the transmission of tactile cues differs depending on the point of contact and the shape of the prosthesis.
Moreover, the mechanical characteristic of a given prosthesis is also expected to affect vibration propagation and, therefore, the intensity and quantity of cues.
The factors we expect to influence this phenomenon include the prosthesis design, the presence of articulations, the rigidity of the assembled mechanisms, and the stiffness of the materials used.
To the best of the authors' knowledge, no study ever investigated how and to what extent these baseline perceptions occur in socket-based prosthetic devices nor provided a quantitative assessment of the roles of the influencing factors.

\begin{figure}
      \centering
      \includegraphics[width=0.85\columnwidth]{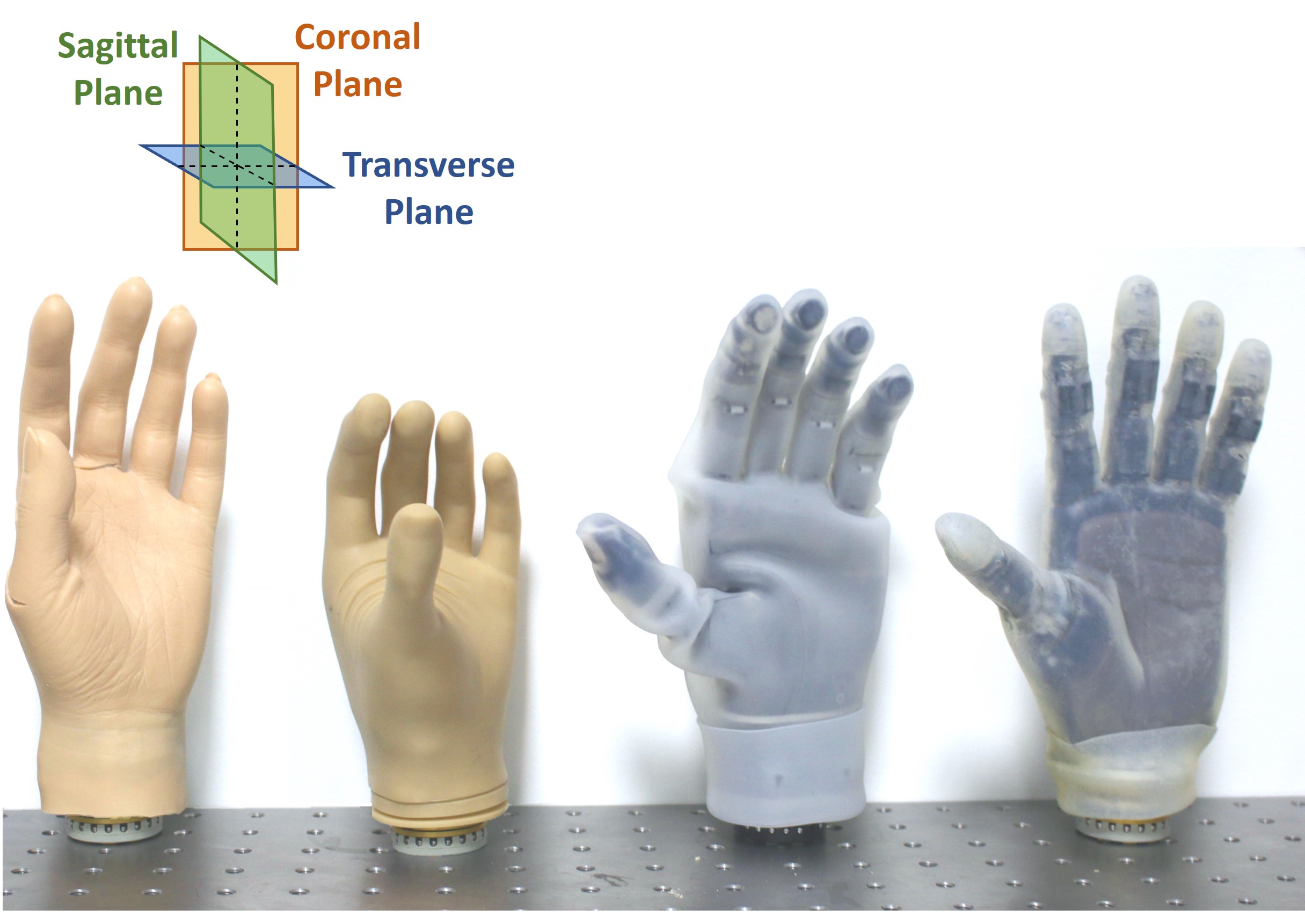}
      \caption{The prosthetic hands adopted in this study: (from left to right) a Cosmetic hand (CH) by Ottobock, a MyoHand VariPlus Speed (VP) by Ottobock, the I-Limb Access (IL) by \"Ossur and a SoftHand Pro (SH). Hands reference planes are shown.}
      \label{fig:hands}
      \vspace{-0.6cm}
\end{figure}

 This study examines the perception and transmission of tactile cues through four prosthetic hands with different characteristics (Fig.~\ref{fig:hands}).
Specifically, we compare hands ranging from rigid, characterized by lower degrees of freedom (DoF), rigid mechanisms, and stiff materials, to advanced ones with more articulations, flexible mechanical designs, and softer materials.
We first report on a pilot study with a prosthetic user, assessing their capability to discriminate which finger is contacted based only on vibration patterns passively transmitted by different hands through the same socket to a user's stump.   
Then, we quantitatively characterize the phenomenon of mechanical vibration transmission of the four prosthetic hands in terms of impact forces and socket-level vibration generated by impacting them with a test machine on the five fingers.
To do so, we design and place a set of vibration sensors in the socket.
Our aim is to correlate those data with user perceptual performance and the different hand designs.
Finally, we examine the information content of the acceleration signals recorded at the stump level to determine whether acceleration data can be used to improve tactile perception and partially compensate for the sensory loss in the softer, more articulated hands.
In other terms, we ask ourselves whether the signals retrieved by the vibration sensors on the socket contain enough information to enable more advanced hands to the same level of discrimination that a user exhibits with rigid hands.
\section{Materials: Prosthetic Hands} \label{sec2}
We tested four bionic hands, representative of the most common mechanical and kinematic characteristics of the currently available prosthetic hands.
The four prosthetic hands selected for the experiments are (Fig.~\ref{fig:hands}):
\begin{itemize}
     \item a Cosmetic hand (CH) by Ottobock\footnote{Cosmetic hand by Ottobock, [Online], Available: \url{https://www.ottobockus.com/prosthetics/upper-limb-prosthetics/solution-overview/custom-silicone-prosthetics/}};
    \item a MyoHand VariPlus Speed (VP) by Ottobock\footnote{MyoHand VariPlus Speed by Ottobock, [Online], Available: \url{https://www.ottobock.com/en-us/product/8E385~59}} (1 DoF);
       \item a I-Limb Access (IL) by \"Ossur\footnote{ I-Limb Access by \"Ossur, [Online], Available: \url{https://www.ossur.com/en-au/prosthetics/arms/i-limb-access}} (5 DoF);
    \item a SoftHand Pro (SH)~\cite{softhandpro} (19 DoF). 
\end{itemize}

The four hands range from rigid designs with fewer DoF, rigid mechanisms, and stiff materials to advanced designs with more articulations, flexible designs, and softer materials.
The Cosmetic hand is a passive hand mainly made of plastic with a silicone glove.
The VariPlus is a tridigit myo-electrically controlled hand with a single actuated degree of freedom opposing a thumb to two fingers, with two passive fingers and a five-finger glove.
The three active fingers (index, middle, thumb) are rigid, made with metallic components, and mechanically coupled (i.e. close simultaneously), and the remaining two follow passively. 
The I-Limb is a myo-electrically controlled hand with metallic components and a soft silicone glove with thicker tips.
The five fingers are individually powered.
Each finger, except for the thumb, has a small mechanical play on the hand sagittal plane (see Fig.~\ref{fig:hands}), while they are rigid in the other directions.
The thumb has a transverse mechanical play.
The SoftHand Pro is mainly made of plastic components with a silicone glove.
A system of elastic ligaments connects the different phalanxes and makes the hand adaptable to all objects, especially in the sagittal hand plane.
At the same time, each finger has a small mechanical play in the coronal plane.
The mechanical structure of the fingers of the four test hands represents four of the main types of finger joints \cite{piazza2019century}: the rigid joint (found in the index, thumb and middle fingers of VP), the continuous joint (in CH, and ring and little fingers of VP), the flexible joint (in IL) and the dislocatable joint (in SH).
\section{Perliminary Analysis}
\begin{figure}
    \centering
    \includegraphics[width=0.95\columnwidth]{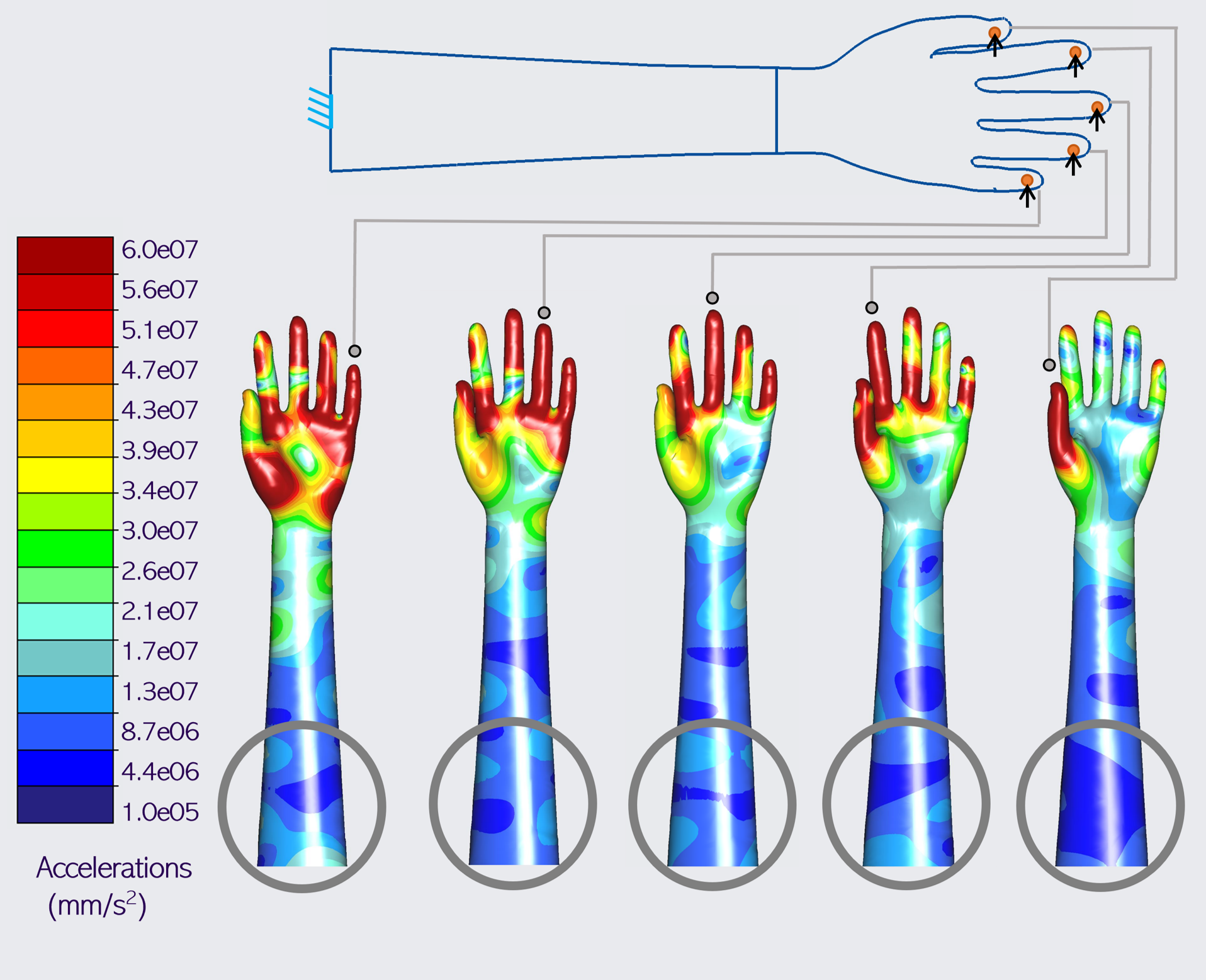}
    \caption{Impact response analysis of a prosthetic-shaped system modelled as a continuous plastic body to an impulse (with a force of 2N) on each fingertip. Acceleration propagation is shown. Grey circles highlight the differences in acceleration propagation depending on the finger contacted. The prosthetic system has a constraint on the end surface of the socket. Simulations are performed on Creo Parametric.}
    \label{fig:first}
    \vspace{-0.6cm}
\end{figure}

We speculate vibrations to play an important role in the natural tactile information transmission through socket-based prosthetic devices.
As a preliminary exploration of this phenomenon, we conduct a response analysis of a simplified generic prosthetic-shaped system. The system is modelled as a continuous body consisting of a socket and hand plastic model ($\rho=\SI{1.33}{\dfrac{kg}{mm^3}},\nu=0.475, E=\SI{32400}{KPa}$), resembling a typical cosmetic prosthesis.
Our investigation revealed that different acceleration patterns arise when a contact cue stimulates different fingers, with each finger experiencing a force of 2N at its respective fingertip (please refer to Fig.\ref{fig:first}).
These accelerations reach the other side of the prosthetic socket and may constitute a crucial component of prosthetic user baseline perception. 
These findings have paved the way for subsequent experiments, motivating our study and prompting further investigation.
\section{Tactile Feedback Perception Experiment} \label{sec3}
We investigate the tactile feedback perception of a prosthetic user.
The ability to discriminate prosthetic fingers during impacts on each fingertip is assessed.
First, a preliminary experiment is conducted with the participant's own cosmetic prosthesis and integrated socket.
Then, four experiments are performed with the four specimen hands mounted on a reference socket.
\subsection{Participant}
One prosthetic user takes part in the experiments and gives their written, informed consent to participate.
The participant (43 years old, female) is affected by limb agenesis and is used to wearing a cosmetic prosthesis even if having experience with myoelectric hands (a SoftHand Pro and a MyoHand VariPlus Speed device). 
The participant's limited experience with the latter devices suggests minimal potential impact on the results. 
The subject has no cognitive impairment that could affect the ability to follow the instructions of the study.
Approval of all ethical and experimental procedures and protocols was granted by the Local Ethics Committee of Area Vasta Nord-Ovest (CEAVNO), Tuscany, Italy, under Protocol No. 7803.
\subsection{Experimental Setup}
\begin{figure}
    \vspace{-0.3cm}
      \centering
      \includegraphics[width=\columnwidth]{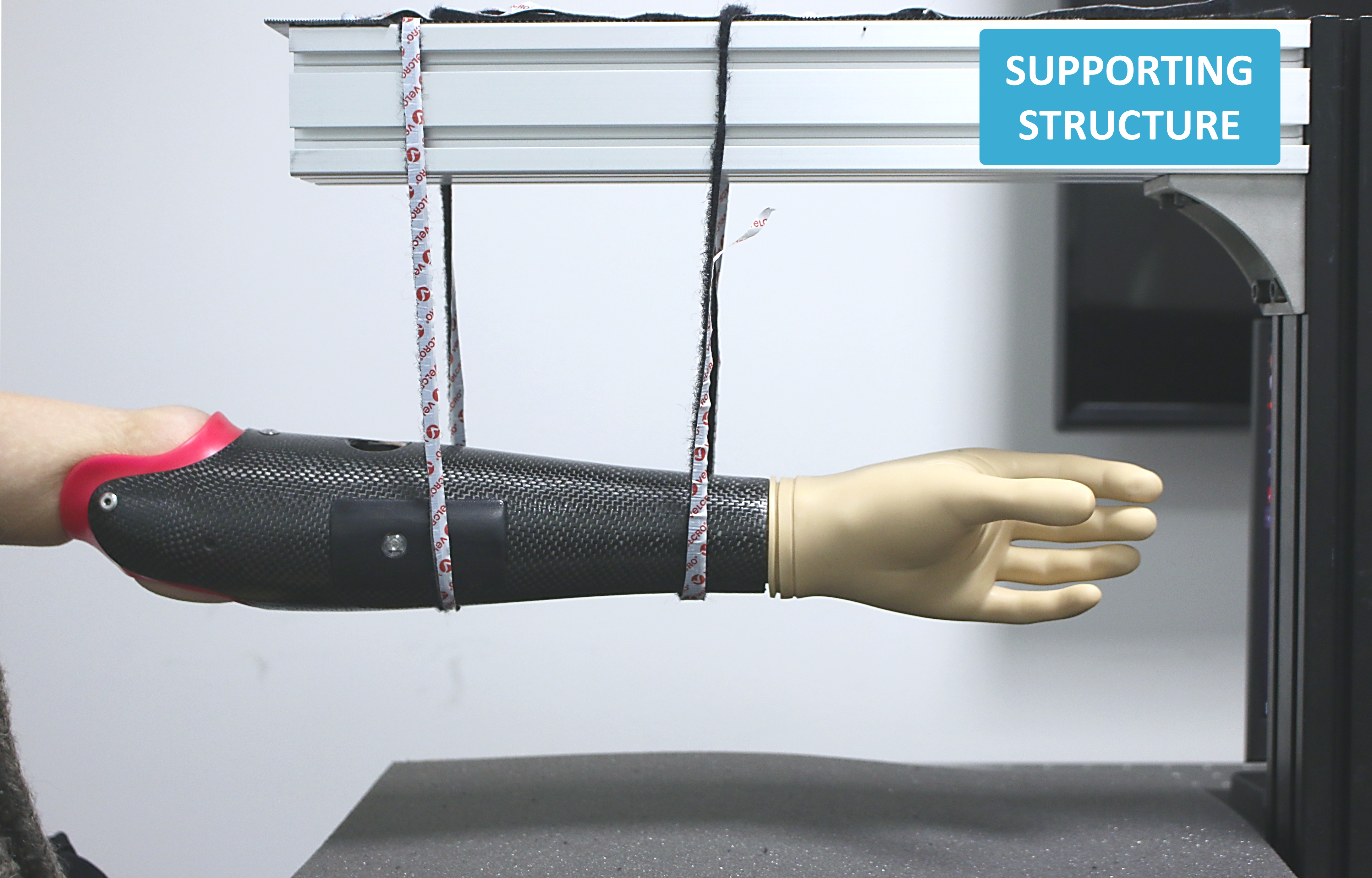}
      \caption{Tactile Feedback Perception Experiment setup:  the prosthetic limb (here with VP hand) is kept in place by a supporting structure made of beams and two cables with Velcro.}
      \label{fig:setuppersonal}
      \vspace{-0.6cm}
\end{figure}

The subject is comfortably seated on a chair in a quiet room.
When the experiment starts, white acoustic noise is provided via headphones to mask any sound from the contacts.
The participant is also blindfolded using a pair of goggles with obscured lenses.
Different positions and supports are tested to select the most comfortable position for the participant and avoid contact transmission being affected by the supporting structure. 
When the arm is freely suspended, the participant experiences discomfort and is unable to maintain a stable arm position throughout the experiments. 
Using a supporting structure for the residual limb prior to the prosthesis device results in compression of the residual limb, which may cause discomfort or pain.
Furthermore, completely supporting the prosthetic arm, such as laying it on a table, compromises impact transmission.
Thus, the final design adopts a supporting structure of loose Velcro loops on which the prosthetic arm is positioned (see Fig.~\ref{fig:setuppersonal}).
By placing the support on the prosthesis instead of a user's skin, the arm position can be kept for all the experiments, and impact absorption is reduced due to the presence of only two Velcro loops.

The same reference socket is employed for all four selected prosthetic hands to ensure that differences arise solely from the variations in hands used. This Ottobock\footnote{Ottobock, [Online], Available: \url{https://www.ottobock.com/en-us/home}} socket, crafted from carbon prepreg for the outer component and silicone for the inner component, was specifically customized for the participant's use. This choice was made to facilitate seamless compatibility with all four prosthetic hands utilized in the study and to replicate the standard socket configuration typically associated with myoelectric prostheses. 
The participant's personal prosthesis consists of a cosmetic hand identical to the Ottobock$^{1}$ CH hand used in this study, without a quick disconnect wrist, and integrated with a plastic and silicone cosmetic socket to create a unified structure. This personal prosthesis does not include any EMG sensors. In the preliminary experiment, the subject's personal cosmetic prosthesis is tested.
Tactile cues are delivered with a small hammer ($\sim\SI{130}{\gram}$) on the prosthesis fingertips, perpendicular to the coronal hand plane (see Fig.\ref{fig:hands}).
The experimenter underwent training to ensure consistent contacts delivery.
The training involved repeated practice sessions to ensure consistent impact delivery to each finger in terms of intensity and direction. Despite diligent efforts, variations in execution were recognized during experimental sessions.
\subsection{Experimental Methods}
A preliminary experiment with the participant's cosmetic prosthesis is carried out to determine the optimal number of impacts for assessing the subject's recognition performance while also ensuring they did not experience undue fatigue or a loss of motivation or concentration.
In order to optimize the subject's comfort, we examined different approaches for delivering impacts, including positioning the subject with their arm suspended or the end of their arm resting on the edge of a table. Ultimately, we select the most suitable position for the user, as depicted in Fig. \ref{fig:setuppersonal}, with the prosthesis hanging from a supporting structure. We then gradually increase the number of impacts per finger, starting with a threshold of 5 impacts. The subject tolerated up to 20 impacts per finger without experiencing discomfort or fatigue.
Based on these findings, we established that delivering 20 impacts to each finger would be appropriate for our experimental purposes.

Each experiment (preliminary experiment included) consists of a \textit{familiarization phase}, a \textit{test phase}, and \textit{rest phase}. 
In the \textit{familiarization phase}, the subject experiences five contacts on each finger without sound and visual insulation.
During the \textit{test phase}, the subject is isolated.
Twenty contacts for each finger are performed with the small hammer.
The order of the finger to be contacted is randomized to prevent the subject from guessing the next one.
After each impact, the participant is asked to answer the question: ``Which finger was contacted?".
The actual finger contacted and the corresponding answer from a user are saved for analysis. 
A \textit{rest phase} of about fifteen minutes is done between one experiment and the other.
The experiment is repeated four times, one for each of the prosthetic hands. 
The order of the specimens is randomized (IL, VP, CH, SH).
The participant is informed that the experiment can be interrupted at any time.
At the end of the experiment, the participant undergoes a subjective evaluation procedure based on a brief questionnaire about prosthetic finger contact perception depending on the prosthetic hand used (Table \ref{tab:quest}).
\begin{table}[]
\centering
\MFUnocap{for}%
\MFUnocap{the}%
\MFUnocap{of}%
\MFUnocap{and}%
\MFUnocap{in}%
\MFUnocap{from}%
\caption{\capitalisewords{ questionnaire.} }
\begin{tabular}{cc} 
\toprule
 \textbf{Questions} &  \\ 
 \hline
 Q1 & \multicolumn{1}{l}{\begin{tabular}[l]{l}Please briefly describe what you feel when each of\\the four prosthetic hands contacts a rigid surface.\end{tabular}}\\ 
 &\\
  Q2 & \multicolumn{1}{l}{\begin{tabular}[l]{l}When the prosthetic hands are contacted, how\\aware are you of which part of the hands, such\\as a finger, has been contacted? Please comment.\end{tabular}}\\
\bottomrule
\end{tabular}
\label{tab:quest}
\vspace{-0.6cm}
\end{table}
\subsection{Data Analysis}
Participant's prediction performance for each hand determines the four scores: $TP$  (True Positive counts), $TN$ (True Negative counts), $FP$  (False Positive counts) and  $FN$ (False Negative counts). 
Those scores are further analyzed using a confusion matrix, and the three metrics accuracy, recall, and precision \cite{grandini2020metrics}. 
Accuracy, defined as: 
\begin{equation}
    Accuracy= \frac{TP + TN}{TP + TN + FP + FN}
\end{equation}
provides an estimation of the correct predictions.
Recall, defined as: 
\begin{equation}
    {Recall= \frac{TP }{TP + FN}}
\end{equation}
tells how frequently the subject can detect a specific category.
Precision measures what percentage of all the positive predictions is truly positive, and it is defined as:
\begin{equation}
    Precision= \frac{TP }{TP + FP}
\end{equation}
All the metrics are compared to the chance level
\begin{equation}
CL=\frac{100\%}{K}=\frac{100\%}{5}=20,
\end{equation}
where $K$ is the number of classes (thumb, index, middle, little, and ring).
The accuracy \cite{grandini2020metrics} considering only the ability to discriminate the thumb finger with respect to the others and the index finger compared to the other fingers are also computed.
This assessment holds significance due to the pivotal roles of these fingers in tasks requiring precise grasping motions.
It underscores the necessity of evaluating their discriminative ability, as it directly impacts the efficiency and precision of prosthetic users in performing daily activities.
\section{Tactile Feedback Transmission Experiment}\label{sec4}
\begin{figure}
      \centering
      \includegraphics[width=0.7\columnwidth]{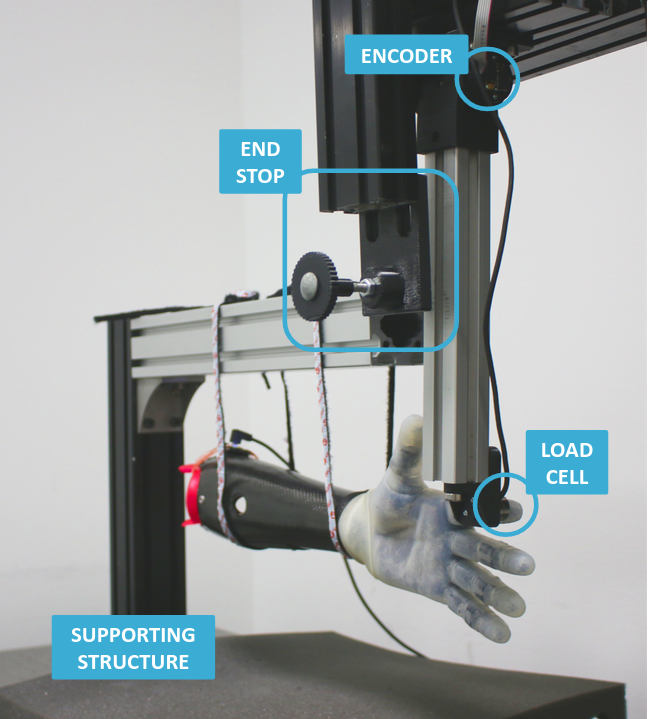}
      \caption{Tactile Feedback Transmission Experiment setup: the pendulum comprises a load cell on the lower extremity, an encoder on the upper extremity, and a custom 3D-printed end-stop to select angles. 
   The Bionic hands (here, the SoftHand Pro) are set with the supporting structure.}
      \label{fig:pend}
      \vspace{-0.3cm}
\end{figure}

\begin{figure}
      \centering
      \includegraphics[width=0.7\columnwidth]{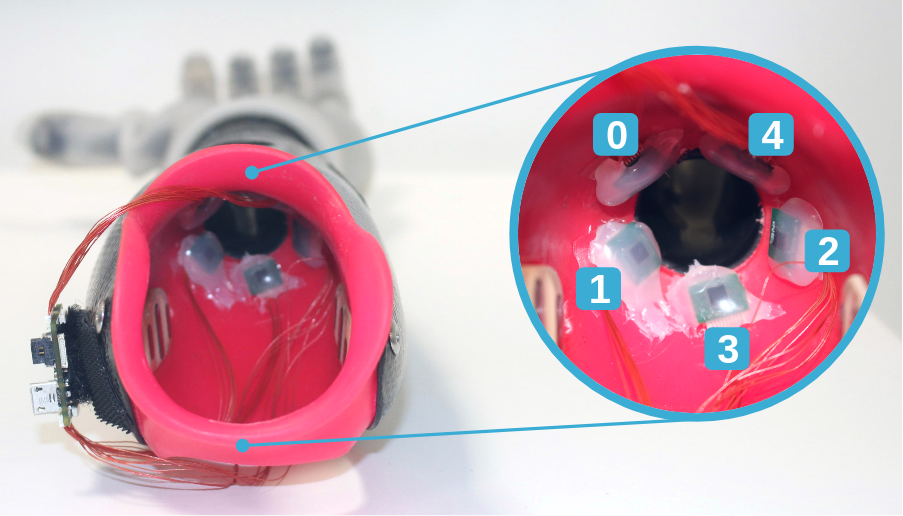}
      \caption{The reference socket is assembled with the I-Limb hand with the IMUs attached inside in radial distribution. A detailed view shows the inner part of the socket and the number of each IMU.}
      \label{fig:imus}
      \vspace{-0.6cm}

\end{figure}

The four hands are investigated to characterize and quantify how the transmission of high-frequency stimuli is affected by the type of prosthesis used.
Each hand's impact responses to quantified contacts are assessed. 
During impacts, the accelerations inside the socket and impact forces of each fingertip are recorded with Inertial Measurement Units (IMUs) and a sensorized pendulum, respectively (see Fig.\ref{fig:pend}, \ref{fig:imus}). 
\subsection{Experimental Setup} 
The four hands are mounted on the reference socket.
The same supporting structure as in Section \ref{sec3} is used.
We attach five IMUs (MPU-9250) to the inner surface of the socket in a radial distribution to measure the differences in vibration patterns after each contact (Fig.~\ref{fig:pend}, \ref{fig:imus}).
This choice is grounded upon the outcomes of the impact analysis performed on a generic prosthetic-shaped system modelled as a continuous body (as shown in Fig.~\ref{fig:first}).
Although the analysis simplifies the system compared to a real prosthetic system, it demonstrates that the same contact on each finger generates distinct acceleration patterns that reach the other side of the prosthesis.
Furthermore, IMUs' position and distribution are also justified by the physical constraints of the residual limb of the participant and its interfacing with the socket.

A sensorized pendulum (see Fig.~\ref{fig:pend}) is used by the experimenter to contact the prosthesis fingertips with repeatable impacts.
As in the first experiment, each contact is perpendicular to the coronal hand plane (Fig.~\ref{fig:hands}).
The pendulum can be moved and adjusted to contact a particular finger in a given position.
Before each impact, the pendulum arm is aligned to an adjustable end-stop to hold it at a desired angle.
Then, the pendulum is released to impact the prosthesis.
During each impact, a load cell (ATI Nano17 Transducer) at the lower extremity of the pendulum measures the impact forces while an encoder (AS5045) at the upper extremity the angles.
\subsection{Experimental Method}
In preliminary tests, the pendulum angle that can create an impact without reaching IMUs full scale is determined and kept for all experiments ($\ang{3}$).
During impacts, acceleration signals inside the socket and impact forces at each fingertip are recorded through electronic boards \cite{NMMI}.
A custom C++ software is developed to synchronize, start, and stop the recordings.
The experiment starts when the pendulum is set at the predetermined angle through the encoder and the end-stop.
The pendulum is left free to swing, and the load cell surface contacts the artificial fingertip.
After each contact, the prosthetic limb oscillates.                    
From the load cell, it is possible to record the three components of the impact force $ F = ( F_x, F_y, F_z )$, while from each IMU inside the socket, the three components of the acceleration $ a = ( a_x, a_y, a_z )$.
The recording session is done at $\sim\SI{1}{kHz}$ to ensure that each impact is properly captured and within the limits of the hardware used to register.
The experiment ends when the pendulum is manually stopped after adequately completing the impact.
Since the recorded signals have proven to be repeatable, three strikes per finger are performed, and one is saved for analysis. This avoids averaging, ensures realistic signals, leverages repeatability, and prioritizes signals without data loss; if no loss occurs, one is chosen randomly.
The experiment is repeated for each robotic hand.
\subsection{Data Analysis} \label{accprocessing}
\begin{figure}
      \centering
      \includegraphics[width=0.85\columnwidth]{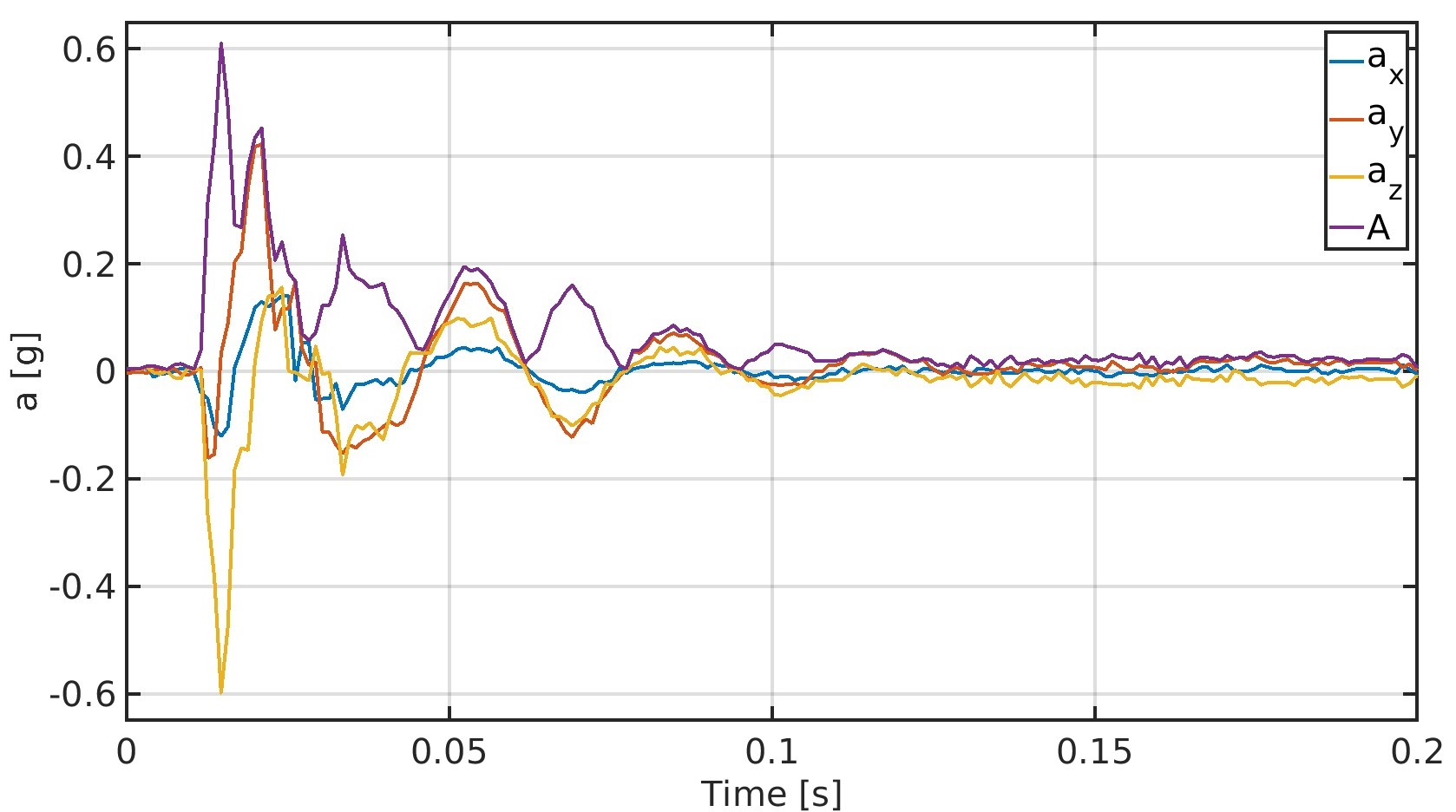}
      \caption{Socket acceleration signal of IMU0 corresponding to a contact on the thumb finger of VP. The three components $a_x, a_y, a_z$ are shown in blue, red, and yellow, respectively and the one-dimensional signal $A$ in purple.}
      \label{fig:acc}
          \vspace{-0.6cm}

\end{figure}

After the experimental session, all signals are analyzed and processed.
The impact force component $F_z$ is selected as the main contribution of the contacts because the pendulum swings in the same direction. 
In regards to the acceleration signals, various processes are carried out to analyze high-frequency signals only.
Communication errors between the electronic boards and the IMUs are removed from the three components $a_x, a_y, a_z$.
Then, a high-pass filter is used to remove the low-frequency oscillation of the prosthesis caused by the contacts. 
Forces and high-frequency accelerations are cut to isolate the contact. 
Each acceleration signal is cut with respect to the impact force peak.
The contact is accurately represented by three hundred samples ($\sim\SI{0.28}{\second}$).
The next step is assigning an energy value to each signal to compare each bionic hand's results quantitatively.
With the algorithm DFT321~\cite{landin2010dimensional}, the three components $a_x, a_y, a_z$ are then reduced into a one-dimensional signal $A$ with the same energy as their sum (Fig.~\ref{fig:acc}).
The DTF321 algorithm, which aims to identify spectral differences, was utilized in our previous works \cite{ivani2023vibes,fani}.
Finally, the acceleration signal energy of each finger is computed for each IMU to compare the results of the prosthetic hands.
The overall mean of each IMU energy signal and fingers is computed for each hand.
Thus, a mean energy value is obtained for the four prosthetic hands. 
To compare results, Spearman rank correlation analysis is computed between the energy mean of each hand and the accuracy of the \textit{Tactile Feedback Perception Experiment}.
The impact force responses of the four hands are also compared.
\section{Tactile Feedback Recognition Experiment}\label{sec5}
To investigate whether acceleration data contain sufficient information to identify which finger is contacted, Artificial Neural Networks (ANNs) are trained.
Due to their capability of handling time series data, Long-Short-Term Memory (LSTM) \cite{hochreiter1997long} models are used for finger detection and trained for each bionic hand.
New acceleration signals inside the socket are recorded upon impacts on each fingertip to create large and balanced datasets.
\subsection{Datasets}
New experiments are carried out to obtain four datasets, one for each robotic hand.
The prosthetic hands with the reference socket are set by the same supporting structure in Section \ref{sec3} and \ref{sec4}.
The five IMUs are kept on the inner surface of the socket to measure the vibration pattern after each contact.
The small hammer is used to contact each fingertip perpendicular to the coronal hand plane (Fig.\ref{fig:hands}).
In preliminary experiments, hammer impacts are tested to determine if the IMUs' full scale is reached.
Then, approximately the same impacts are manually delivered in all experiments.
The experiments consist of one hundred impacts on each fingertip.
The three acceleration components $ a = ( a_x, a_y, a_z )$ are recorded from the five IMUs inside the socket during each impact. 
Then, the same data processing in Section \ref{accprocessing} is carried out to obtain the one-dimensional signal $A$.
Thus, four datasets are obtained with five hundred samples each, one hundred for each fingertip. 
Each sample comprises five one-dimensional signals $A$ from the five IMUs.
\subsection{LSTM Network Architecture and Hyperparameters}
As the purpose of this experiment is to determine whether finger contact information can be extracted from acceleration data, various types of classification algorithms may be appropriate.
LSTM networks are a particular type of Recurrent Neural Network (RNN) able to process entire data sequences and selectively remember information, widely used to capture time correlations efficiently \cite{staudemeyer2019understanding,hochreiter1997long}.
To avoid a separate feature extraction step, we choose LSTM over linear support vector machines (SVMs).
We train four LSTM models to recognize the finger contacted from the acceleration signals $A$ of each IMU.
Each dataset is randomly split into 80\% training, 10\% validation and 10\% testing sets. 
We experimentally test different network parameters and hidden layers to evaluate the best performance.
The four nets are composed of a first feed-forward layer with ReLU activation function and a second normalization layer which normalizes all five features together (the five signals from the IMUs).
Then, the third layer is an LSTM layer.
The output is fed to a dropout layer.
The four layers repeat, and, lastly, the softmax layer gives the probability of each class (thumb, index, middle, ring, little), and the classification layer computes the cross-entropy loss.
We set the dropout to 0.2.
The hyperparameters are tuned by using the Bayesian optimization algorithm in MATLAB (MathWorks Inc., Natick, MA, USA), maximizing the validation accuracy on the validation sets.
The hyperparameters selected are the initial learning rate, the number of units in a dense layer, the number of hidden units, the number of epochs, and the batch size.
\subsection{Data Analysis}
Test sets are used to evaluate the performance of the four networks.
Four confusion matrices with True Class and Predicted Class are computed. 
The results are compared in terms of accuracy, recall and precision metrics \cite{grandini2020metrics}.
\section{Results}\label{sec6}
\subsection{Tactile Feedback Perception Experiment}
We assessed the finger discrimination ability of the subject wearing CH, VP, IL and SH bionic hands.
The subject's cosmetic hand was also tested in the preliminary experiment.
With the subject's own prosthesis, the accuracy and mean recall reached $55\%$ and $58\%$, respectively.
Results of the experiments with the four specimens are shown in Table \ref{tab:perceptionresults}.
The accuracy related to the CH hand is $58\%$ while VP and IL accuracies are $52\%$ and $45\%$, respectively.
With SH, the accuracy reaches $37\%$.
The subject was able to discriminate the finger contacted with an accuracy that is always above the chance level.
Mean precision and mean recall were also computed for each hand. 
Since the same number of contacts were delivered for each finger, the results are balanced, and the test accuracy also corresponds to the mean recall \cite{grandini2020metrics}.
\begin{table*}[]
\centering
\MFUnocap{for}%
\MFUnocap{the}%
\MFUnocap{of}%
\MFUnocap{and}%
\MFUnocap{in}%
\caption{\capitalisewords{Subject accuracy and mean precision of the \textit{Tactile Feedback Perception Experiment}  for each bionic hand.}}
\begin{tabular}{lcclccl}
\multicolumn{1}{c}{\textbf{Bionic   Hands}} &
  \textbf{Accuracy{[}\%{]}} &
  \multicolumn{2}{c}{\textbf{Mean Precision{[}\%{]}}} &
  {\color[HTML]{000000} \textbf{\begin{tabular}[c]{@{}c@{}}Thumb   Recognition \\ Accuracy{[}\%{]}\end{tabular}}} &
  \multicolumn{2}{c}{\textbf{\begin{tabular}[c]{@{}c@{}}Index   Recognition\\ Accuracy{[}\%{]}\end{tabular}}} \\ \hline
\textbf{Cosmetic}  & 58 & \multicolumn{2}{c}{62} & 81 & \multicolumn{2}{c}{83} \\
\textbf{VariPlus}  & 52 & \multicolumn{2}{c}{63} & 82 & \multicolumn{2}{c}{65} \\
\textbf{I-Limb}    & 45 & \multicolumn{2}{c}{49} & 93 & \multicolumn{2}{c}{72} \\
\textbf{Soft-Hand} & 37 & \multicolumn{2}{c}{45} & 61 & \multicolumn{2}{c}{55} \\ \hline
\end{tabular}%
\label{tab:perceptionresults}
\end{table*}

All the mean precision results are higher than the mean recall metrics. 

If it's considered only the ability to discriminate the thumb finger with respect to the others, positive results are obtained for the accuracy \cite{grandini2020metrics} of each hand: $81\%$ for CH, $82\%$ for VP, and $93\%$ and $61\%$ for IL and SH, respectively.
Regarding index recognition accuracy compared to the other fingers, the results are: $83\%$ for CH, $65\%$ for VP, $72\%$ for IL and $55\%$ for SH.

\begin{figure*}
\centering
\includegraphics[width=.45\textwidth]{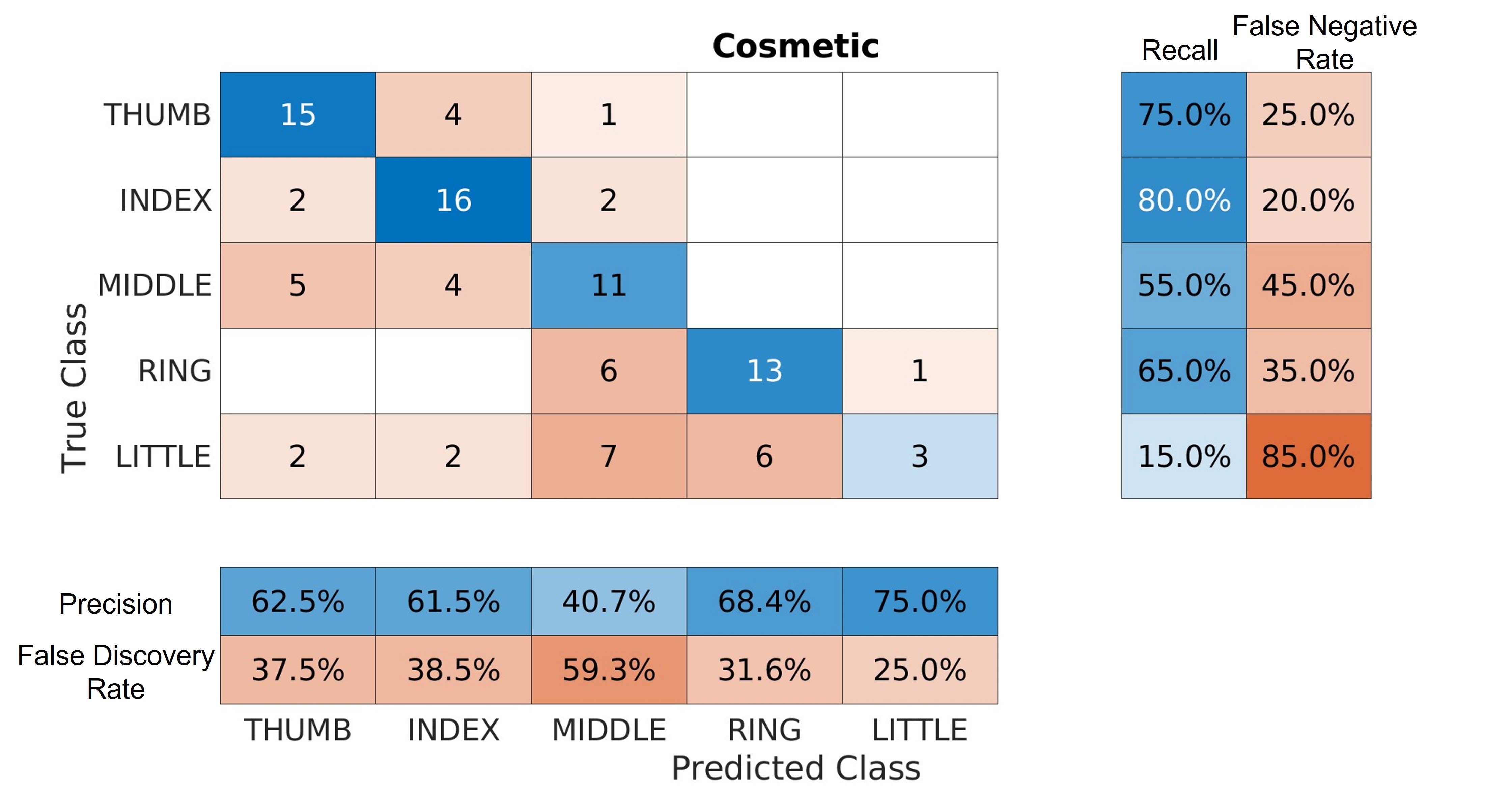}\quad
\includegraphics[width=.45\textwidth]{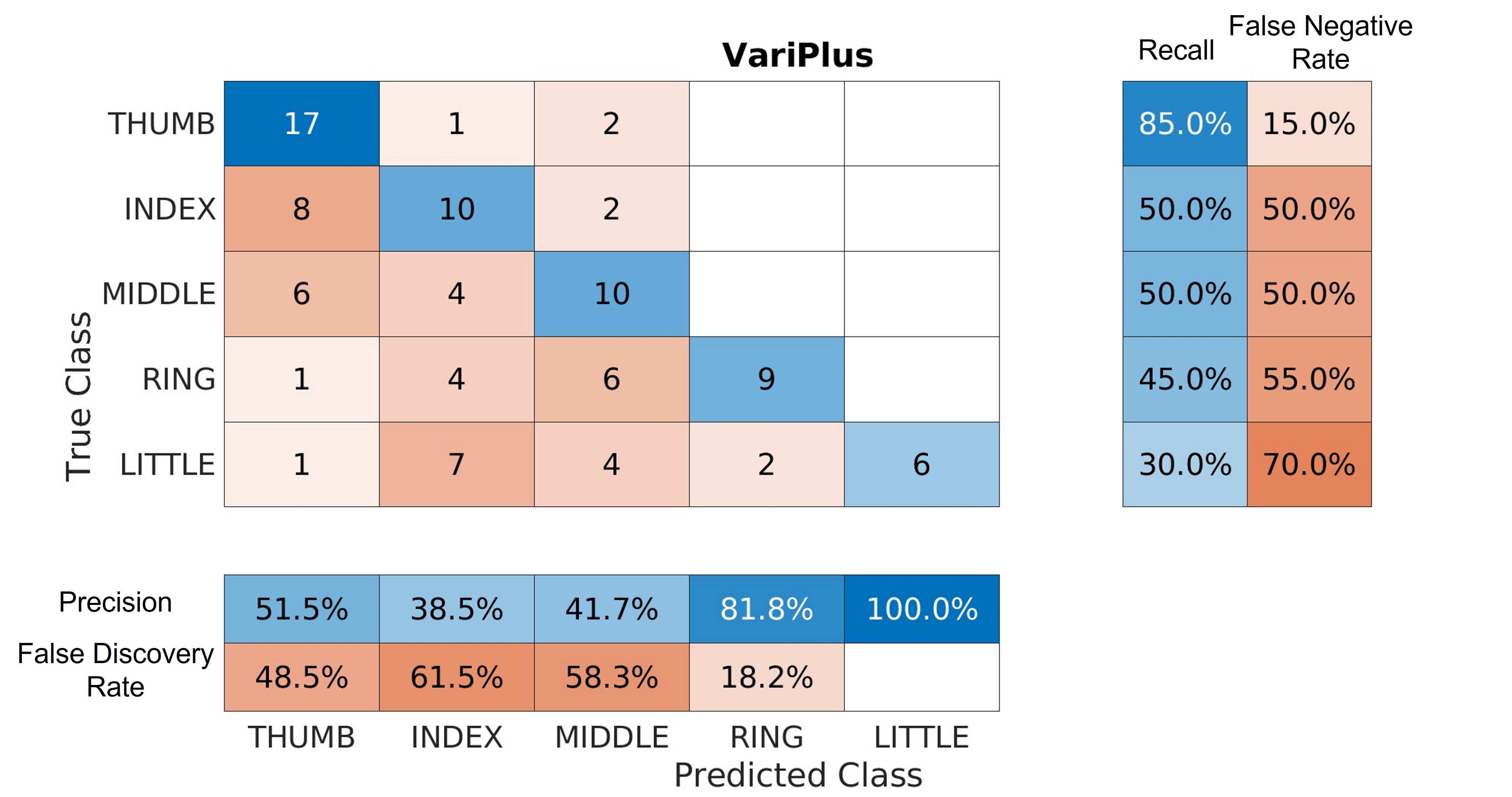}
\medskip
\includegraphics[width=.45\textwidth]{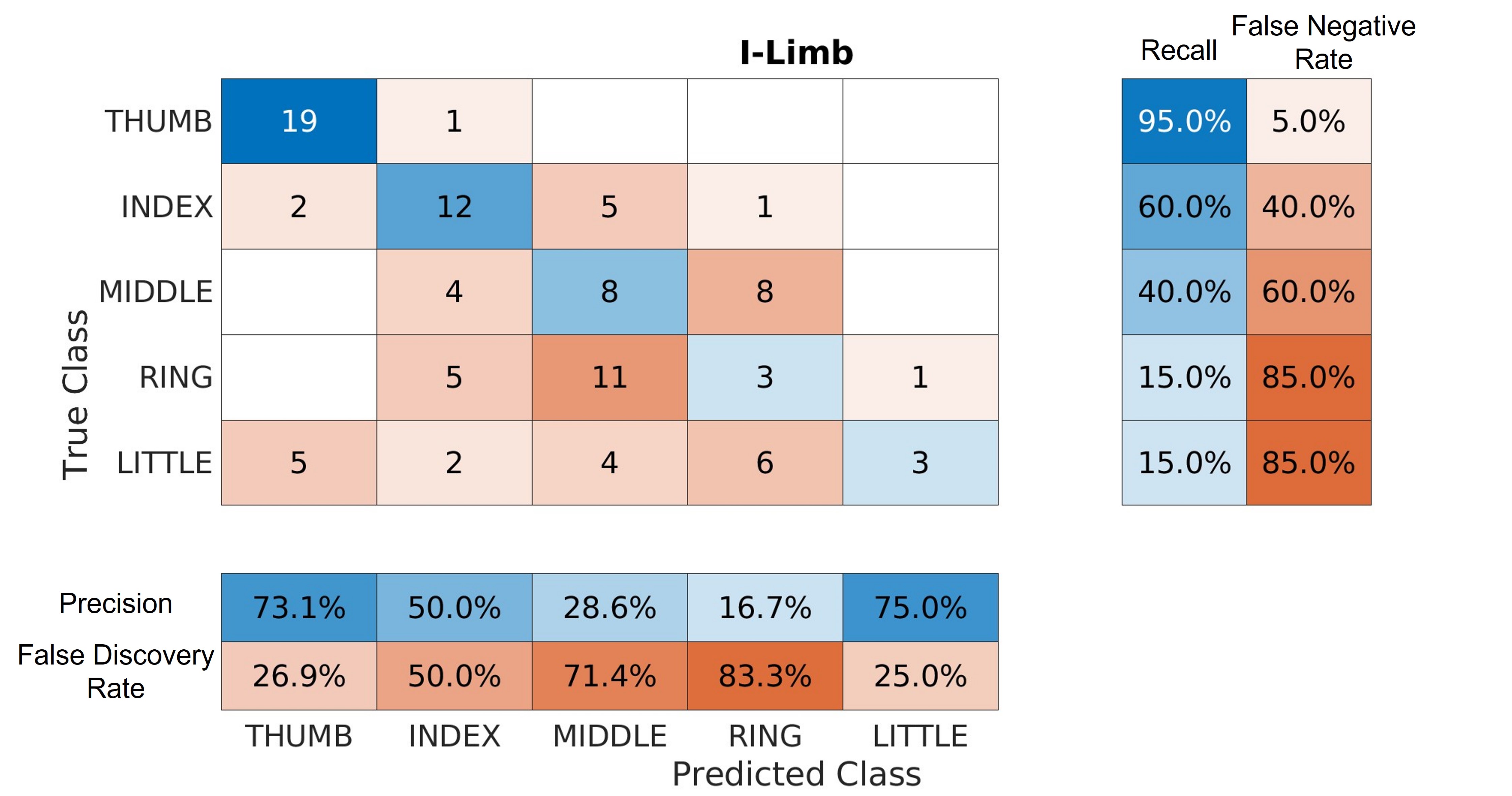}\quad
\includegraphics[width=.45\textwidth]{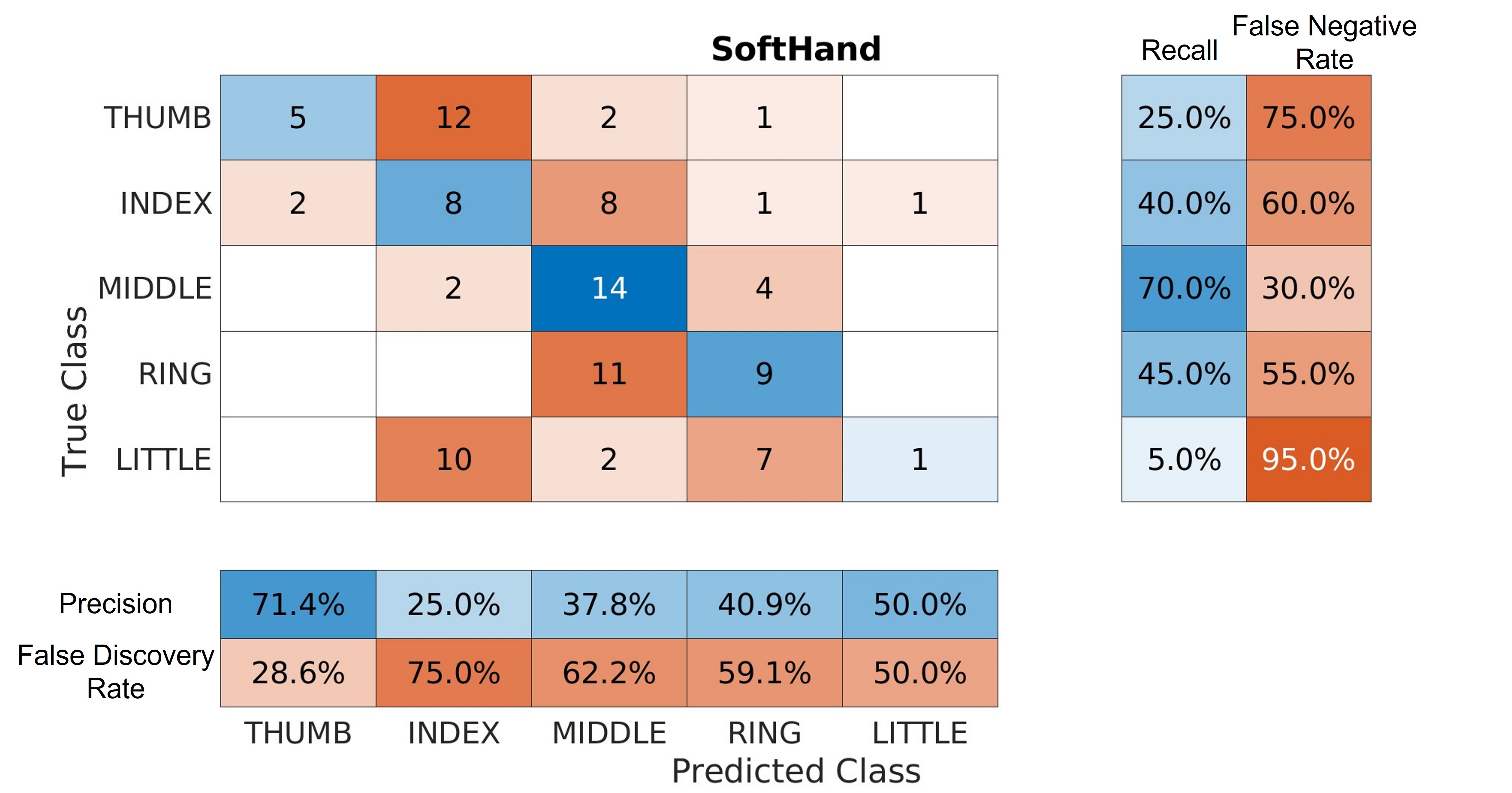}\quad
\caption{Tactile Feedback Perception Experiment: Confusion Matrices of the results with CH, VP, IL, and SH bionic hands of finger discrimination experiments. The row and column summary displays the percentage of correctly classified and incorrectly classified observations for each true or predicted class. The first column summary is the recall metric of each class, while the first row summary is the precision metric of each class.}
\label{fig:perceptionconfusionmat}
    \vspace{-0.3cm}

\end{figure*}

In Fig.~\ref{fig:perceptionconfusionmat}, confusion matrices of the four bionic hands are shown.
Each row summary and column summary displays the percentage of correctly classified and incorrectly classified observations for each true or predicted class.
In particular, the first column summary is the recall metric of each class, while the first row summary is the precision metric of each class.  
About the VP hand, higher accuracy and recall are highlighted for the thumb, the index and the middle fingers, which are more rigid with respect to the other fingers. 
This could be due to the presence of the three metallic fingers under the glove.
A $95\%$ of recall and a $73\%$ of precision for the thumb of IL hand reflects the lower mechanical play in that direction of impact with respect to the other fingers.
With the SoftHand, the overall discrimination ability of the subject decreased. 
The worst recall metric was achieved with the little finger in all hands, while precision is quite high for the same finger. 
Thus, the subject was rarely able to detect the little finger, but most of the predictions were correct.  

Based on the findings from the questionnaire (please refer to Table \ref{tab:quest_answ}), the subject perceived vibrations within the inner socket of the prosthesis. Notably, these vibrations were reported to be smoother when using the SH and more discernible when utilizing the CH. Additionally, the subject could attribute the source of these vibrations to a specific contact zone.
\begin{table}[]
\centering
\MFUnocap{for}%
\MFUnocap{the}%
\MFUnocap{of}%
\MFUnocap{and}%
\MFUnocap{in}%
\MFUnocap{from}%
\caption{\capitalisewords{ questionnaire Answers.} }
\begin{tabular}{cc} 
\toprule
 \textbf{Answers} &  \\ 
 \hline
 Q1 & \multicolumn{1}{l}{\begin{tabular}[l]{l}I don't feel big differences between all rigid hands.\\ The huge difference is between the SH and the others.\\In all rigid hands, I feel the vibration on the inner socket,\\ while with the SH I feel a smoothed vibration.\end{tabular}}\\ 
 &\\
  Q2 & \multicolumn{1}{l}{\begin{tabular}[l]{l}The more the hand/socket are rigid, the more I can \\ clearly feel the zone/finger contacted. The hand that\\ I can feel more precisely is the CH.\\ With that kind of hand, I really can feel each finger.\\ The worst case is the SH. Its softness decreases a lot the\\ vibration and, as a consequence, the richness of the\\ sensation that arrives in the inner socket.\end{tabular}}\\
\bottomrule
\end{tabular}
\label{tab:quest_answ}
    \vspace{-0.6cm}

\end{table}
\subsection{Tactile Feedback Transmission Experiment}
\begin{figure}
      \centering
      \includegraphics[width=0.94\columnwidth]{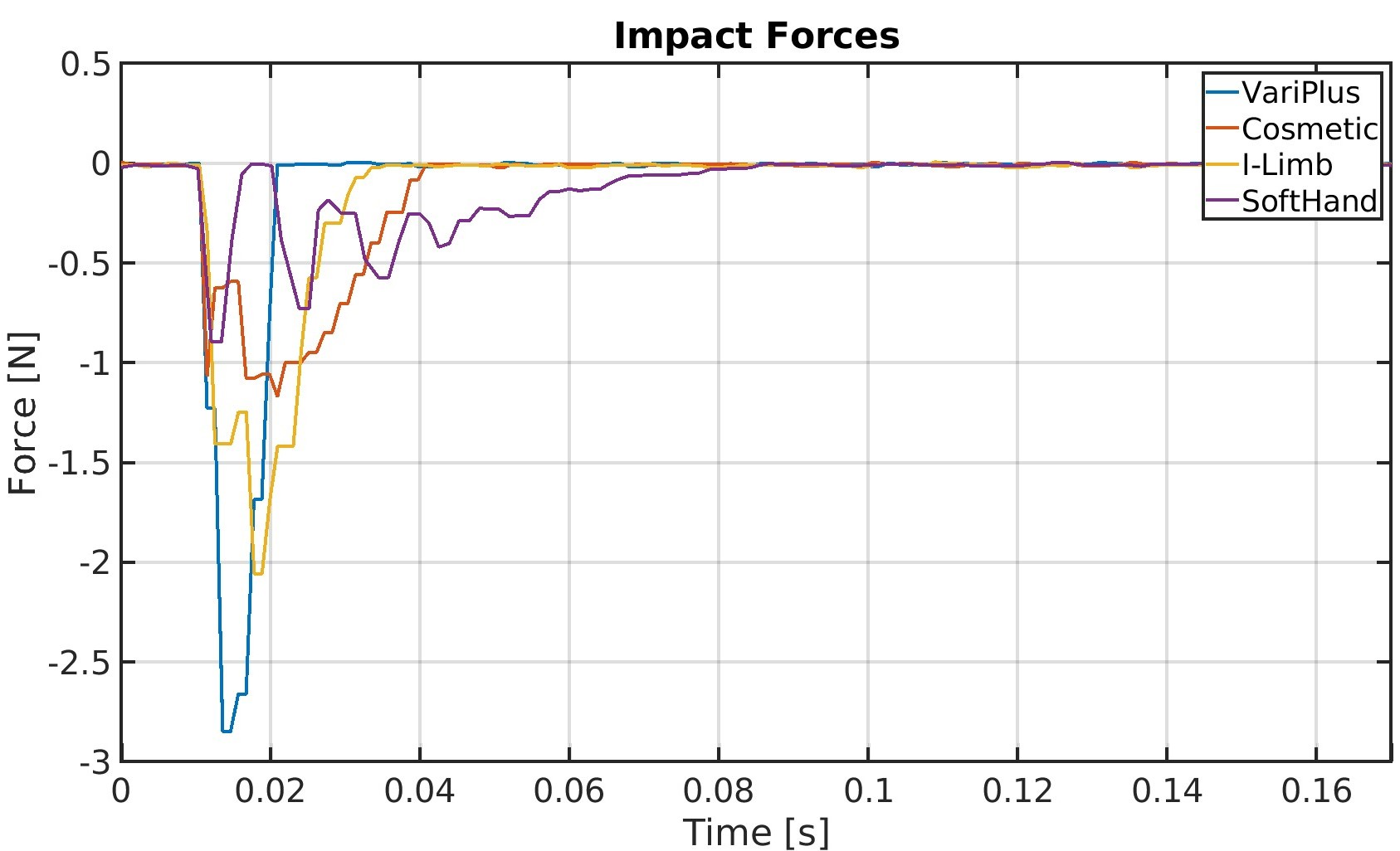}
      \caption{Tactile Feedback Transmission Experiment: Impact forces of the sensorized pendulum with the index finger for VP, CH, IL and SH.}
      \label{fig:forces}
      \vspace{-0.6cm}
\end{figure}

We recorded the acceleration signals from IMUs glued inside the socket and the impact forces of the pendulum, which contacted the prosthetic fingertips.
Fig.~\ref{fig:forces} shows the different impact forces of the sensorized pendulum with the index finger of each bionic hand.
It can be noticed that the impact forces of the index finger of less compliant hands (as VP) are intense and shorter in time.
In such a case, the impact between the pendulum and the prosthesis is almost instantaneous, and the energy is, thus, quickly transmitted.
With more compliant hands, the response is less intense and slowly transmitted. 
    
\begin{figure*}
    \vspace{-0.5cm}

\includegraphics[width=\columnwidth]{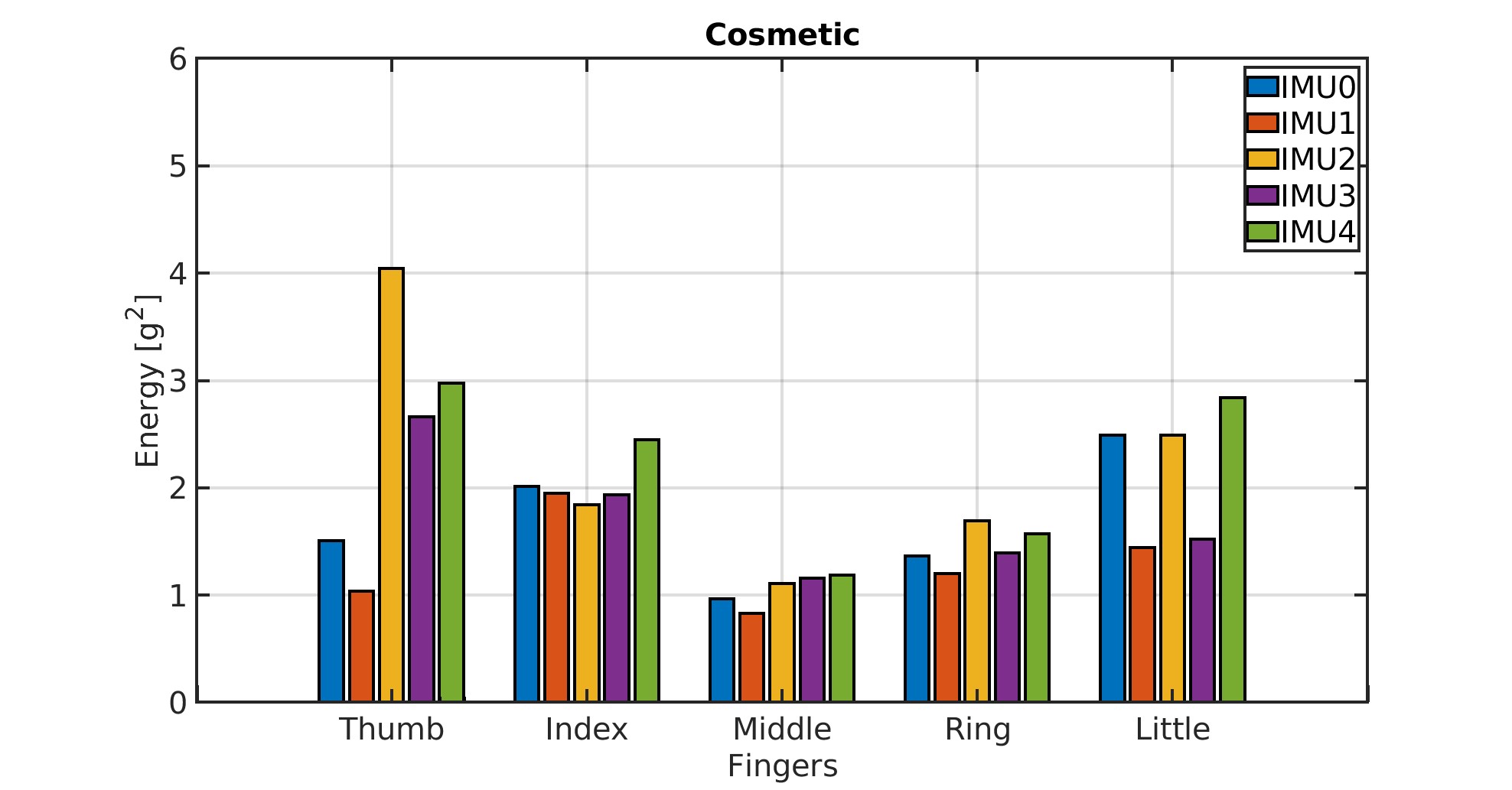}\quad
\includegraphics[width=\columnwidth]{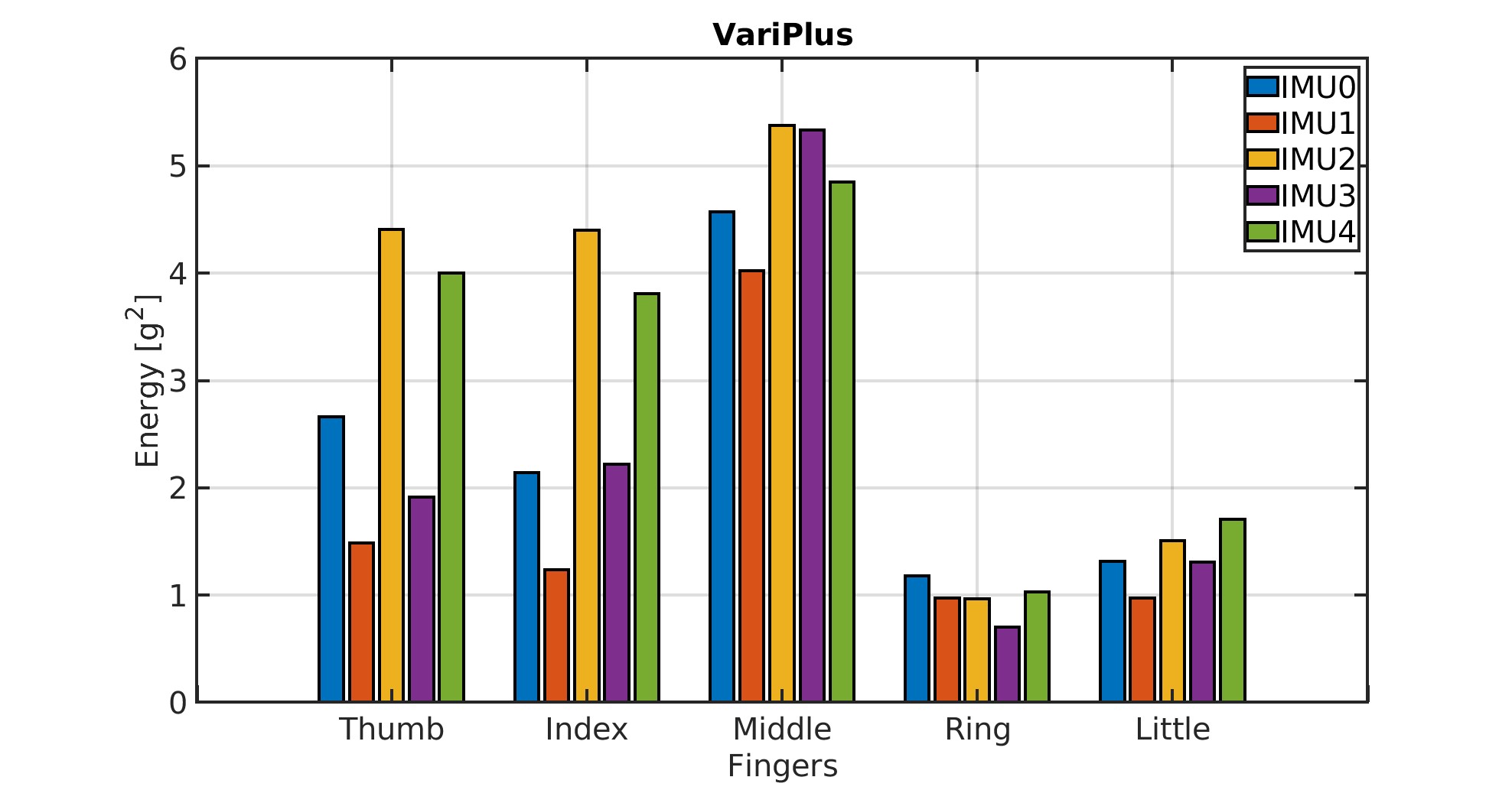}\medskip

\includegraphics[width=\columnwidth]{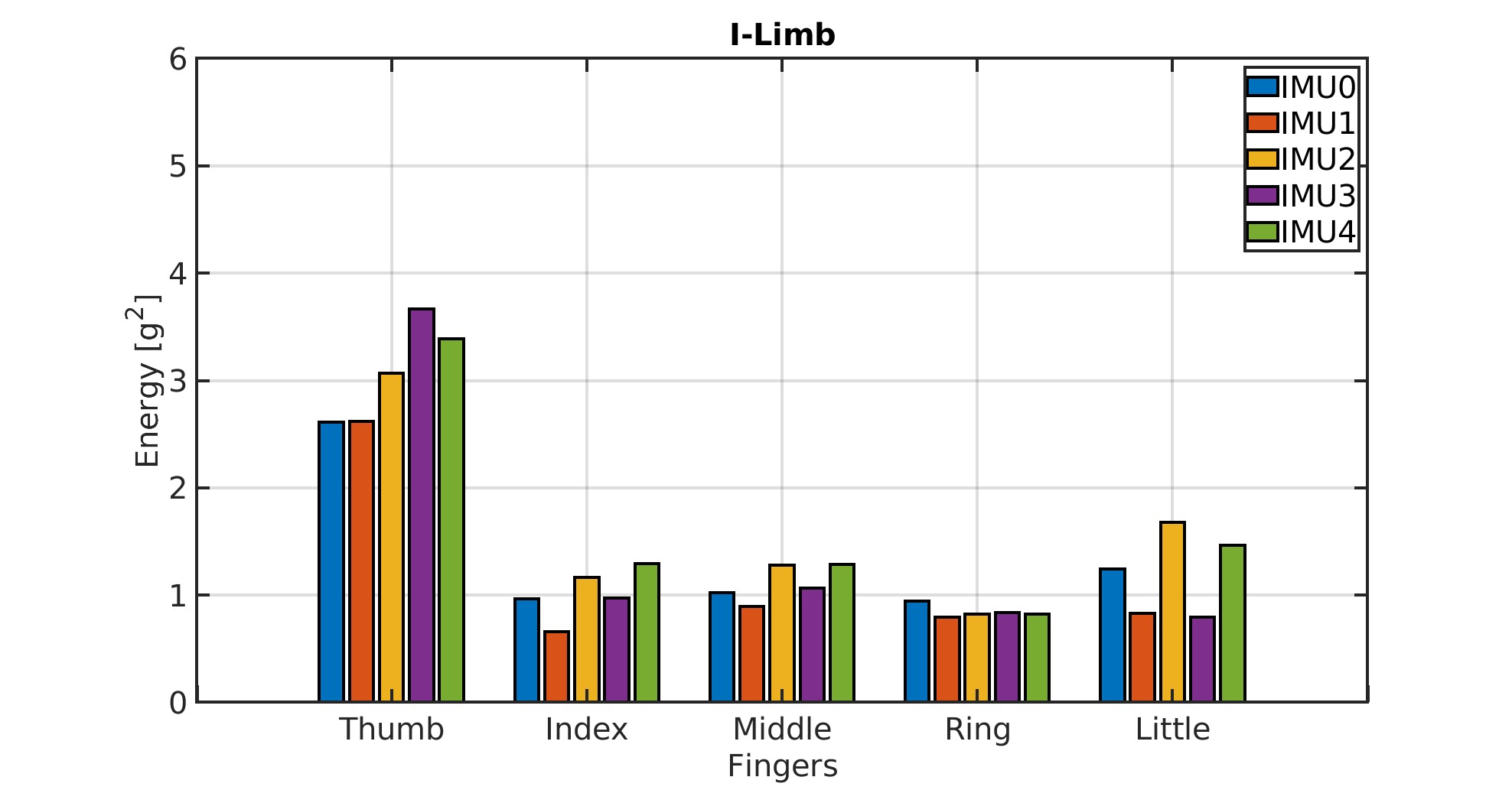}\quad
\includegraphics[width=\columnwidth]{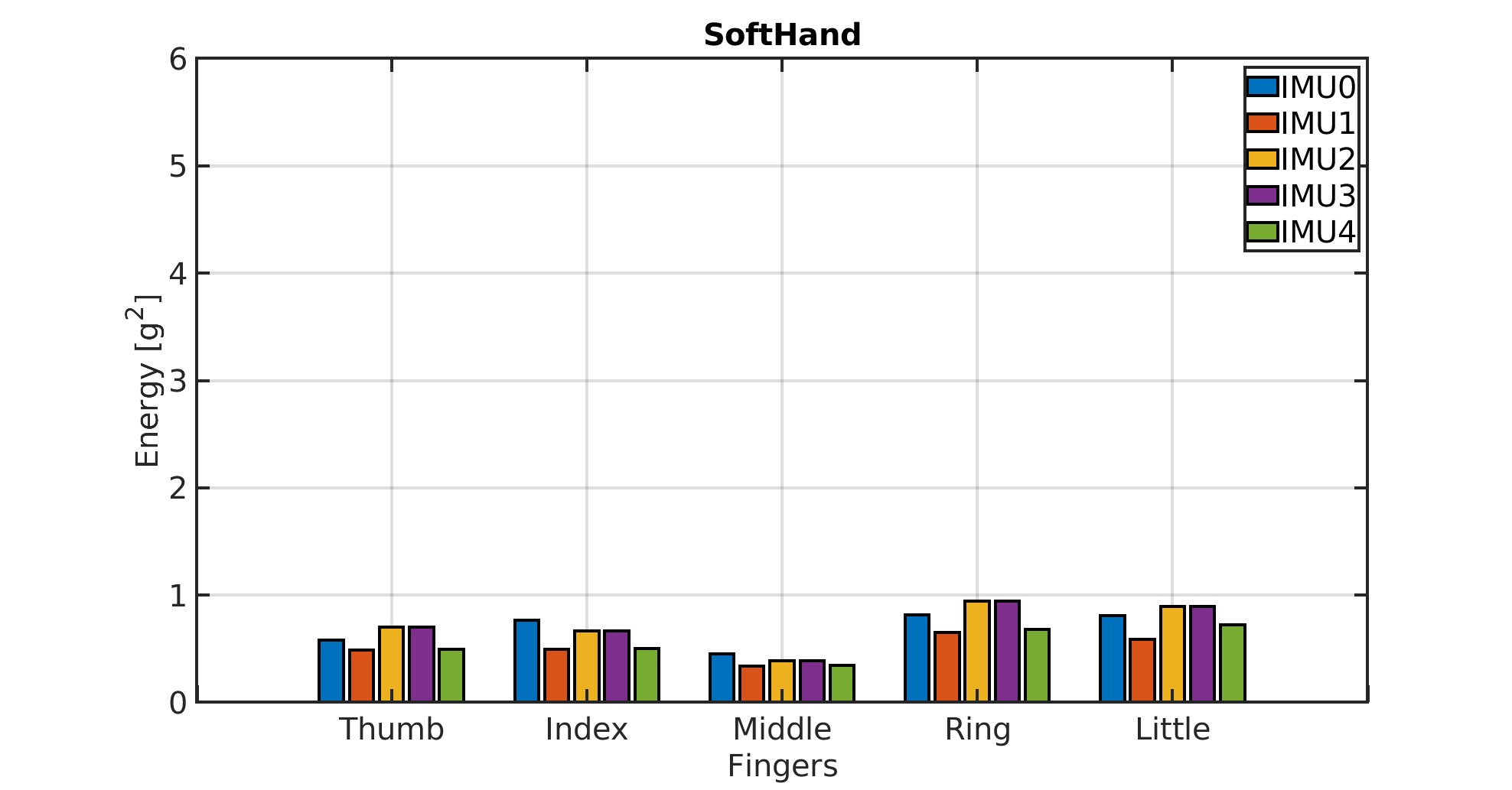}\quad
\caption{Tactile Feedback Transmission Experiment: Bar charts of the computed energy of each $A$ signal from the five IMUs inside the socket after an impact on each fingertip. Each chart corresponds to  CH, VP, IL, SH hands. IMUs' numbers refer to Fig.~\ref{fig:imus}.}
\label{fig:energybartable}
   \vspace{-0.015cm}

\end{figure*}

The bar charts in Fig.~\ref{fig:energybartable} show the computed energy of the one-dimensional signals $A$ for each IMU. 
IMUs' numbers refer to Fig.~\ref{fig:imus}.
Since the socket is always the same, transmission differences depend on the type of hand used.
The energy associated with socket accelerations is higher when the rigid VariPlus and Cosmetic hands are worn (especially in the metallic fingers of VP) and decreases with the other hands.
Indeed, high-frequency signals are better transmitted in more rigid designs.
Since the IL thumb is stiffer in the direction of impact, its energy results are greater than the other fingers.
For the softer and more articulated hands (SH and IL), the energy is much lower for each IMU.
In SH, IMUs energy values are also similar between a finger and another.
Indeed, it has several dampening elements, such as the wrist, which prevent vibration transmission through the prosthetic device.

The behaviour of the five IMUs in response to contacts is slightly different from each other in all fingers.
This can be noticed especially in the more rigid and less articulated fingers.
The Spearman rank coefficient between the mean energy value of each hand and the accuracy of \textit{Tactile Feedback Perception Experiment} was 0.8.
Thus, a strong monotonic association was found.
\subsection{Tactile Feedback Recognition Experiment}
We trained four LSTM networks to recognize the fingers contacted based on acceleration data recorded inside the socket, one net for each bionic hand. 
Nets' hyperparameters were tuned using Bayesian optimization.
Table \ref{table:hyper1} shows the hyperparameters tuning results, and Table \ref{table:netresults} shows validation and test accuracy results and mean precision metrics for the prosthetic hand used. 
\begin{table}[]
\centering
\MFUnocap{for}%
\MFUnocap{the}%
\MFUnocap{of}%
\MFUnocap{and}%
\MFUnocap{in}%
\MFUnocap{from}%
\caption{\capitalisewords{Nets' hyperparameters resulted from tuning the four training and validation sets in the \textit{Tactile Feedback Recognition Experiment}.} }
\begin{tabular}{ccccc} 
\toprule
 \textbf{Hyperparameters} & \textbf{LSTM$_{VP}$}& \textbf{LSTM$_{CH}$} & \textbf{LSTM$_{IL}$} & \textbf{LSTM$_{SH}$}  \\ 
 \hline
\multicolumn{1}{l}{\textbf{\begin{tabular}[l]{c}Initial\\Learning Rate\end{tabular}}} & 0.0011 & 0.0016  & 0.0027 & 0.0033\\
 \textbf{Epochs} & 68 & 168 & 185 & 82\\
\multicolumn{1}{c}{\textbf{\begin{tabular}[]{c}Units in \\a Dense Layer\end{tabular}}} & 40  & 37 & 21 & 18\\
 \textbf{Hidden Units} & 40 & 40  & 39 & 39\\
 \textbf{Batch Size} & 64 &  32 & 64 & 64 \\
\bottomrule
\end{tabular}
\label{table:hyper1}
    \vspace{-0.1cm}

\end{table}
Since all the datasets were balanced, the test accuracy results also correspond to the mean recall \cite{grandini2020metrics}.
\begin{table}[]
\centering
\MFUnocap{for}%
\MFUnocap{the}%
\MFUnocap{of}%
\MFUnocap{and}%
\MFUnocap{in}%
\MFUnocap{from}%
\caption{\capitalisewords{Nets' validation and test accuracy in the \textit{Tactile Feedback Recognition Experiment}.}}
\begin{tabular}{cccc}
\toprule
\textbf{Net} &
  \multicolumn{1}{c}{\textbf{\begin{tabular}{c}Validation\\Accuracy{[}\%{]}\end{tabular}}} &
  \multicolumn{1}{l}{\textbf{\begin{tabular}[l]{c}Test\\Accuracy{[}\%{]}\end{tabular}}} &
  \multicolumn{1}{l}{\textbf{\begin{tabular}[l]{c}Test Mean\\Precision{[}\%{]}\end{tabular}}} \\
  \hline
 \textbf{LSTM$_{CH}$} & 84 & 78 & 79  \\
 \textbf{LSTM$_{VP}$} & 96 & 86 & 86 \\
 \textbf{LSTM$_{IL}$} & 88 & 88 & 89 \\ 
\textbf{LSTM$_{SH}$} & 76 & 78 & 78  \\ 
\bottomrule
\end{tabular}
 \label{table:netresults}
     \vspace{-0.4cm}

\end{table}
All networks were able to detect the finger contacted with positive results. 
Validation and test accuracy of the LSTM model trained on the CH socket acceleration data reached $84\%$ and $78\%$, while on VP data reached $96\%$ and $86\%$, respectively.
For SH, the validation and test accuracy are $76\%$, $78\%$, and for the IL, they both are $88\%$.
Mean precision and recall are also balanced.
Accuracy, recall and precision metrics are all higher than the \textit{Tactile Feedback Perception Experiment} results.

\begin{figure*}
   \vspace{-0.44cm}
\centering
\includegraphics[width=.45\textwidth]{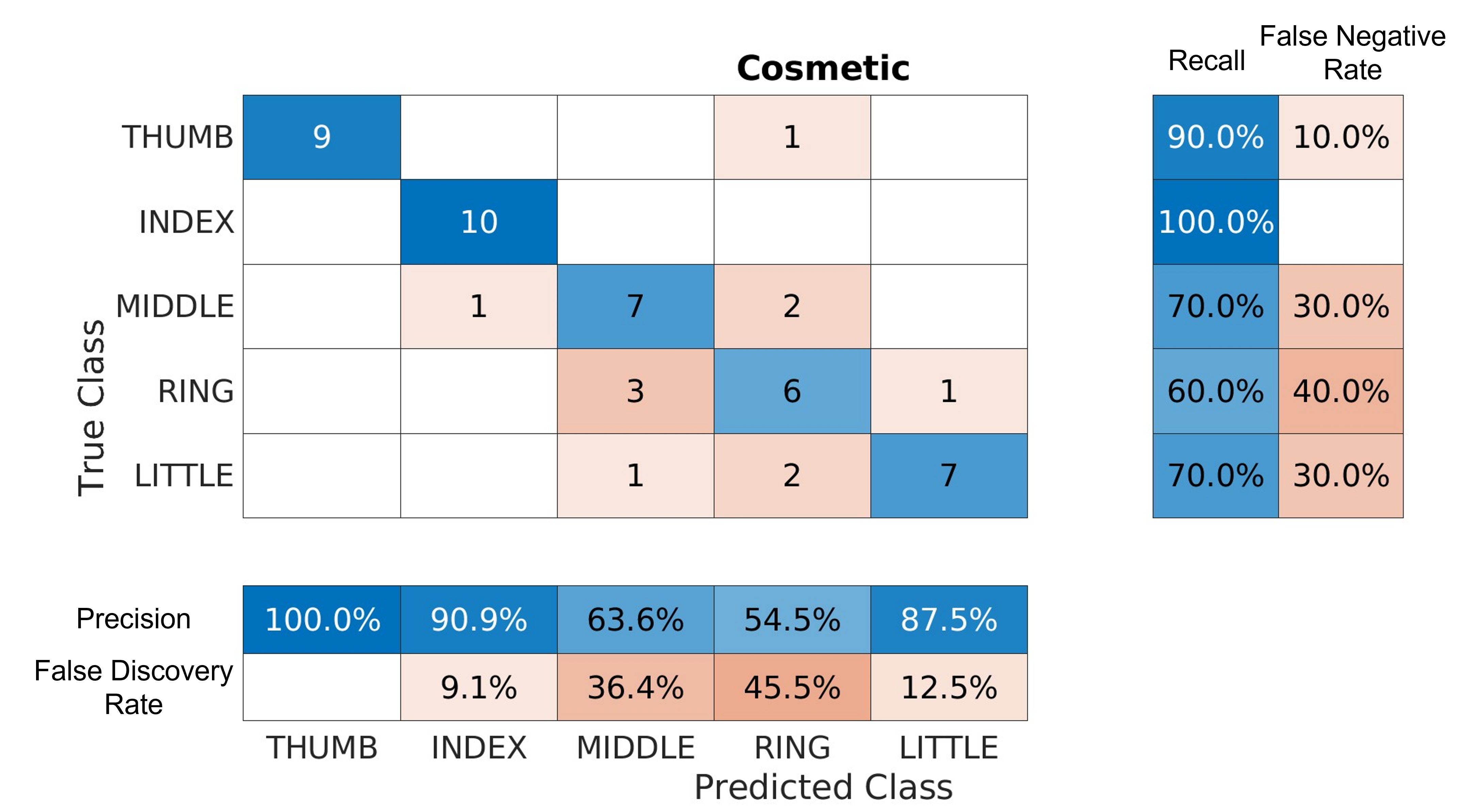}\quad
\includegraphics[width=.45\textwidth]{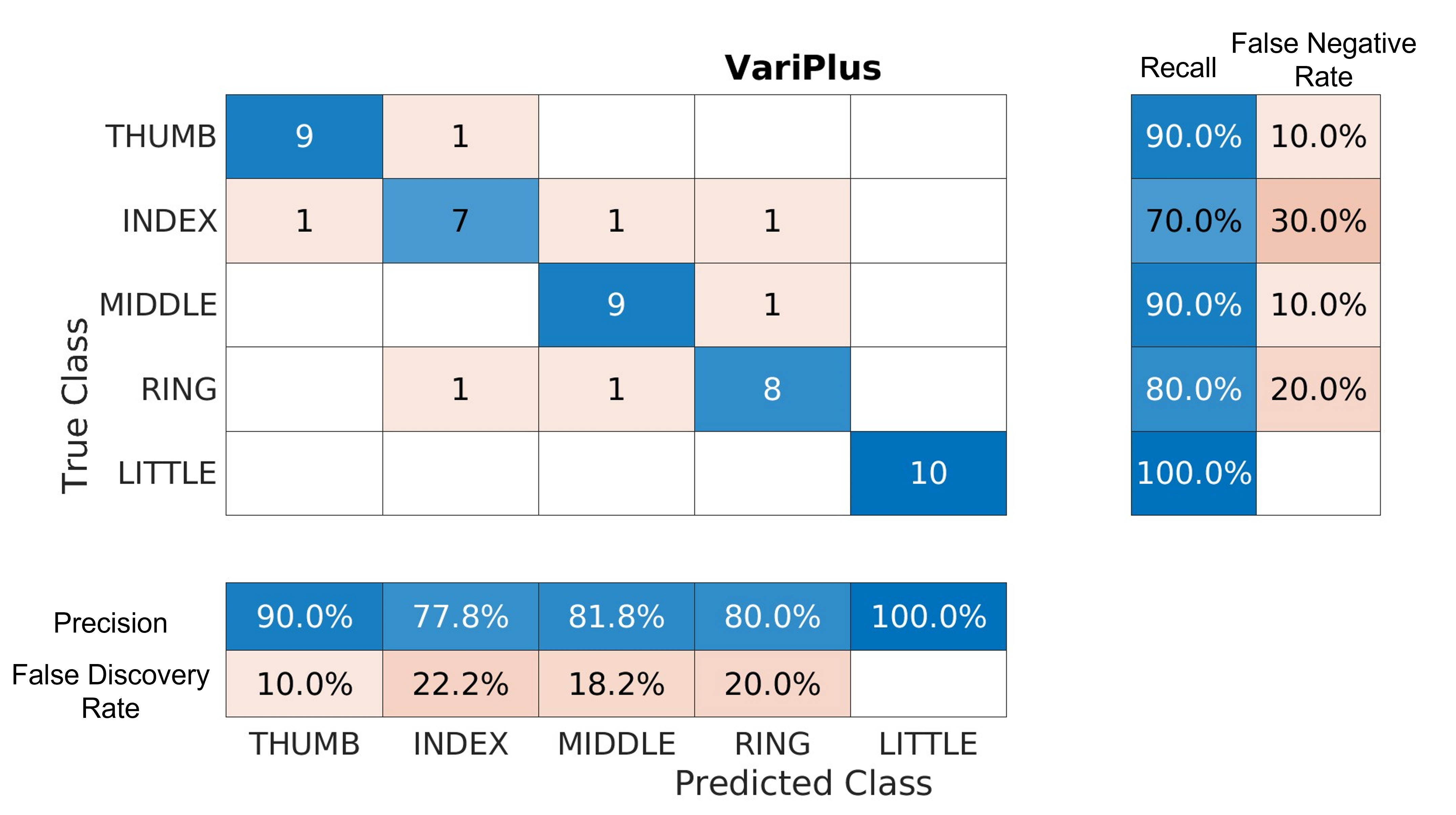}
\medskip
\includegraphics[width=.45\textwidth]{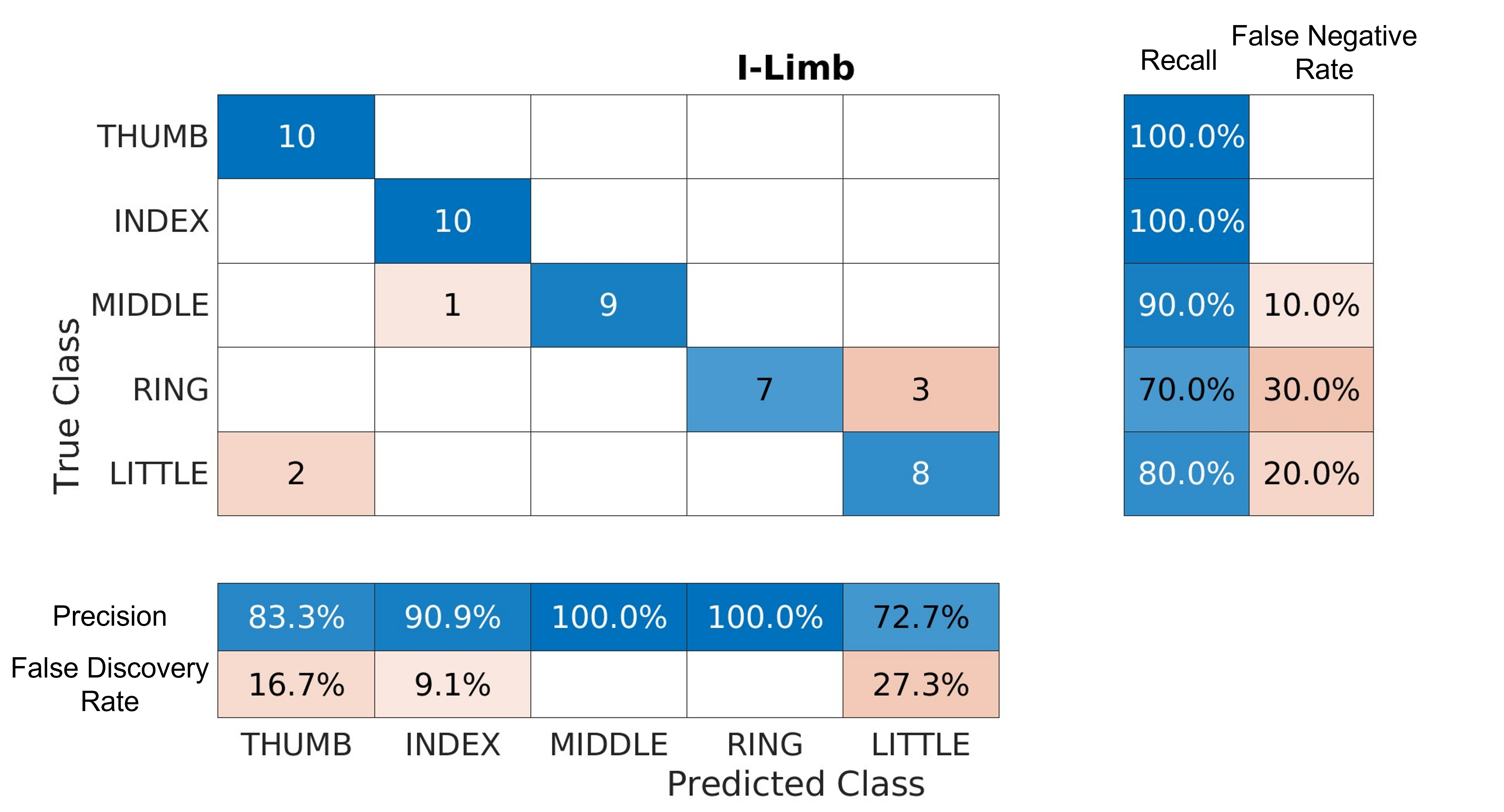}\quad
\includegraphics[width=.45\textwidth]{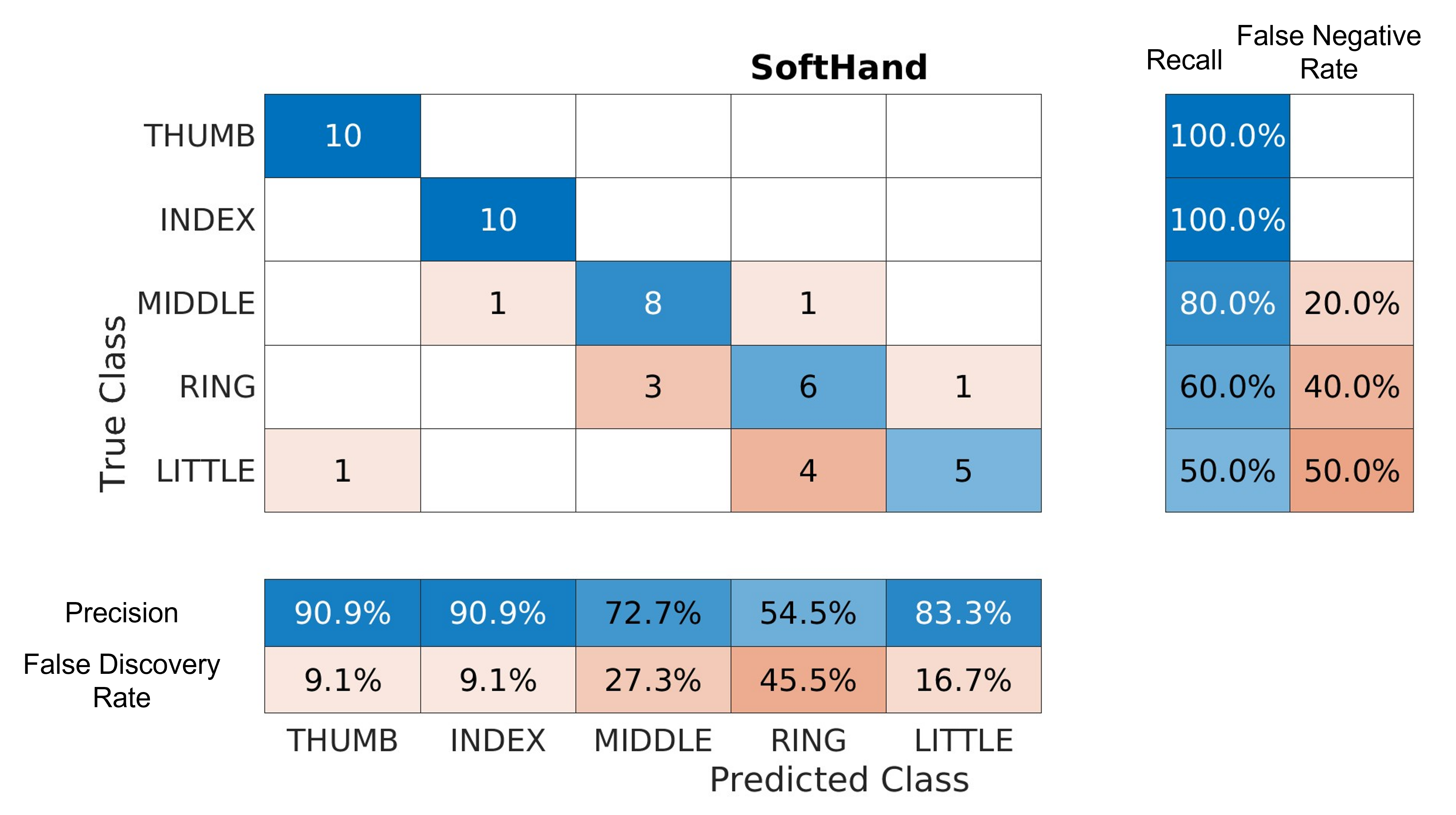}\quad
\caption{Tactile Feedback Recognition Experiment: Confusion matrices of the results of CH, VP, IL, and SH LSTM nets evaluated on test sets. The row and column summaries show the percentage of correctly and incorrectly classified observations for each class. The first column provides the recall metric for each class, while the first row provides the precision metric.}
\label{fig:recognitionconfumat}
    \vspace{-0.6cm}

\end{figure*}

Confusion matrices of the four LSTM networks evaluated on test sets are shown in Fig.~\ref{fig:recognitionconfumat}.
From the confusion matrices, it should be noted that each class was well predicted from all LSTM models.
Precision and recall metrics are also balanced for each class.
The overall performance decreased for the SoftHand LSTM model.
\section{Discussion}\label{sec7}
\subsection{ \textit{Tactile Feedback Perception Experiment}}
    \vspace{-0.1cm}
The \textit{Tactile Feedback Perception Experiment} proved that the subject was able to perceive and recognize the contact on all the bionic hands.
The subject detected the finger contacted with positive results.
Indeed, the accuracy related to all the bionic hands is much higher than the chance level of 20\%.
The subject's personal prosthesis used in the preliminary experiment includes a CH hand, the same as that used in the experiments, along with a cosmetic socket. Despite the subject being used to their cosmetic prosthesis, which they don regularly, better accuracy was observed with the CH hand and the reference socket in the experiment with respect to that measured in the preliminary one. This difference could be due to the tighter and more rigid structure of the reference socket used in the experiments compared to the user's own cosmetic socket.
Thus, vibrations may be better transmitted to the subject.

Results prove that the subject could associate the transmitted vibration patterns to a finger position with better outcomes for rigid and less articulated hands.
Future studies will investigate the potential learning curve of this skill.
Wearing the Cosmetic and the VariPlus hands, accuracy metrics are 58\% and 52\% with respect to 45\% and 37\% of the I-Limb and the SoftHand bionic hands.
Regarding the index finger recognition compared to the other fingers, the subject achieved impressive results, with higher accuracy for the CH hand (83\%) than VP and IL (65\% and 72\% accuracies, respectively). 
With the SH, only a 55\% accuracy was achieved, slightly higher than the chance level.
Considering the discrimination ability to identify a contact on the thumb from an impact on the other fingers, the higher recognition accuracy for the I-Limb (93\%) compared to the CH (81\%) and the VP(82\%) can be explained by considering that the I-limb thumb has a transverse mechanical play. 
Thus, during the impact, set on the hand sagittal plane, the thumb is more rigid compared to the other fingers and may make it easier for the subject to detect high-frequency vibration associated with that finger compared to the CH and VP. 
The SoftHand decreases the performance (accuracy of 61\%), but still with a result above the chance level.

The findings from the questionnaire analysis offer valuable insights into the tactile feedback perception within prosthetic devices. 
The discernible differences in vibration perception between SH and CH could potentially be attributed to various factors, including the material properties of the sockets and the manner in which vibrations are transmitted through the hands. 
The more discernible perception with CH indicates a more efficient transmission of vibrations, which may result in reduced sensory fatigue for the user. 
Moreover, the participant was able to consciously localize the source of vibrations to specific contact zones, underscoring the importance of taking this vibration transmission into account for sensory feedback device design in prosthetic devices. 
This localization capability may facilitate more intuitive interaction with the prosthesis, allowing users to adapt their movements and behaviours accordingly.
Overall, these findings highlight the intricate interplay between design factors and user perception in shaping the efficacy and usability of prosthetic devices.
\subsection{\textit{Tactile Feedback Transmission Experiment}}
    \vspace{-0.1cm}
In the \textit{Tactile Feedback Transmission Experiment}, we have effectively quantified impacts at the socket level and studied the response of each hand.
The results quantitatively reinforce our intuition that advanced, softer, and more articulated hands dampen contacts, while rigid hands with stiff components and fewer articulations better transmit high-frequency vibrations.
Indeed, more intense accelerations were detected at the socket level for the more rigid hands with respect to the advanced ones.
While the observed outcome aligns with expectations, it's crucial to note the lack of prior empirical evidence directly supporting this hypothesis. 
The findings reflect the \textit{Tactile Feedback Perception Experiment} results despite being carried out under different setup conditions.
In the presence of IMUs, the subject could not wear the prosthesis, and the vibration transmission behaviour is different with respect to when the arm is present.
Nevertheless, hanging the prosthesis enabled us to characterize each robotic hand and understand how vibration is transmitted regardless of the person wearing the prosthesis. 
The Spearman rank coefficient of 0.8 demonstrates that user perception and measured accelerations inside the socket are monotonically correlated.
The hands with lower energy socket vibrations are those with lower accuracy results and vice versa.  
\vspace{-0.1cm}
\subsection{ \textit{Tactile Feedback Recognition Experiment}}
    \vspace{-0.1cm}
In the \textit{Tactile Feedback Recognition Experiment}, we proved that LSTM models are able to detect the finger contacted from the high-frequency socket acceleration signals with quite a high accuracy for all hands.
Thus, the recorded vibratory patterns contain the information necessary to discriminate the impacted finger, even if registered on the inner surface of the socket.
The discrimination ability of ANNs is substantially better than the subject's one in all prosthetic hands. 
The distinct data collection methods in this experiment compared to the \textit{Tactile Feedback Transmission Experiment} are driven by their divergent objectives. 
In the \textit{Tactile Feedback Transmission Experiment}, the use of the sensorized pendulum facilitated consistent quantification of each impact, allowing for analysis and comparison across different hands. Conversely, in the \textit{Tactile Feedback Recognition Experiment}, we aimed to replicate the conditions of the \textit{Tactile Feedback Perception Experiment}, necessitating the introduction of differences between impacts using a small hammer rather than producing identical impacts.

For acceleration dimensional reduction, we used the DTF321 algorithm in both \textit{Tactile Feedback Recognition Experiment} and \textit{Tactile Feedback Transmission Experiment}. 
As mentioned, the DTF321 algorithm is considered one of the best 321 approaches for offline processing when perceptual similarity is prioritized \cite{9773011}. 
Still, DTF321 is not rotation invariant, and a different orientation of the IMU yields different DFT321 values even if the signals captured by the IMU are identical in the new orientation of the sensor \cite{landin2010dimensional,9773011}.
Thus, future online implementation will consider the Principal Component Analysis (PCA) algorithm to avoid orientation dependencies~\cite{9773011,8986575}.
Nevertheless, as a result of our findings, the recorded acceleration signals and ANNs have the potential to compensate for the loss of tactile perception given by the advanced hands.
\subsection{Limitations}
\vspace{-0.1cm}
About the \textit{Tactile Feedback Perception Experiment}, we acknowledge the limitation arising from the inclusion of only one prosthetic user in the experiment.
Obviously, the findings are subjective, emphasizing the necessity for future studies to validate them with an increased and heterogeneous group of participants, encompassing individuals with varying nature, levels, location of amputations and state of the nerves in the residual limb, which can affect the user’s perception.
Moreover, it is important to consider that the participant's perception may be influenced by external variables, such as their state of mind, discomfort, or boredom, alongside factors related to skin condition, thus adding complexity to the interpretation of the results.
Furthermore, it is essential to acknowledge that our assessment of prosthetic user performance was conducted under controlled conditions, specifically in a quiet room, where the participant was fully engaged in the designated task. This deliberate choice was made to minimize extraneous factors that could potentially detract from the accuracy of our findings, although at the expense of mirroring real-world situations. Our future research will encompass experiments conducted in environments that more accurately replicate real-life scenarios.

Deliberately choosing a small hammer over the sensorized pendulum used in the \textit{Tactile Feedback Transmission Experiment} aimed to avoid significant delays in setup and repositioning for each contact. This decision was made to maintain the subject concentration and minimize fatigue during the experimental session. Additionally, it allowed for replicating realistic impact scenarios like those encountered in everyday environments and suggested that the ANNs are also able to discriminate among different levels of impact. 
However, experimenter errors may have prevented consistent impact levels and directions for each finger and repetition.
Future research will explore alternative methods to deliver contact in a more quantified manner while still prioritizing the avoidance of fatigue or discomfort for the subject.
Furthermore, our research will explore various types of prosthetic impacts and directions.
Exploring changes in impact location, orientation, amplitude, and contact size on the prosthetic hand could yield valuable insights in future studies.
We also aim to examine how simultaneous impacts interact to produce sustained and continuously modulated cues and whether similar identifiable propagation patterns emerge during shear loading.

The choice of experimental materials, notably the type and material of the socket, influenced our findings. It is reasonable to posit that variations in these factors may have significantly impacted the transmission of vibrations through the socket, thereby shaping the perceptual responses of the subjects involved in our experiments. This assertion is substantiated by our study, which demonstrated the influence of different prosthetic hands on perception.  Future work will be focused on conducting additional tests aimed at systematically investigating the effects of diverse socket types on both vibration transmission and subject perception. This will necessitate a comprehensive exploration of various socket designs and materials to gain deeper insights into their influence on the overall performance of prosthetic devices.
Nevertheless, by consistently employing the same reference socket throughout our experiments, we were able to effectively discern differences among the various prosthetic hands used.

The presence of inertial measurement units (IMUs) prevented the subject from wearing the prosthesis, potentially leading to improved classification accuracies in the Tactile Feedback Recognition Experiment. However, this setup simplified post-impact oscillation filtering by using a high-pass filter to remove low-frequency socket oscillations.  Indeed, the oscillations from the suspended socket differ greatly in frequency from the high-frequency oscillations associated with the vibration modes of the prosthesis. Any link to a rigid support, like the human arm, would accelerate the damping of the high-frequency oscillations via contact transmission, complicating measurement. However, it's crucial to note that these transmitted vibrations are precisely what elicit human perception, aligning with our measurement objective. 
Addressing the potential alteration of prosthesis dynamics in the absence of a human arm, future research will explore methods for measuring impulse propagation while a prosthetic user wears the socket, ensuring comfort is not compromised.
    \vspace{-0.2cm}
\subsection{Study Significance and Future Directions}
    \vspace{-0.1cm}
Our results indicate that part of the tactile information in exploring the external world is transmitted through the prosthesis and is not lost, with differences based on the type of hand.  
In rigid prosthetic designs, the transmitted information has allowed the subject to associate vibration transmission with specific tactile cues. Consequently, this may diminish the necessity of haptic feedback devices (e.g. vibrotactile actuators for transmitting initial contact or texture cues). 
However, in the case of softer and more articulated prosthetic designs, this correlation is not consistently observed.
Although prosthetic hands with soft and compliant components are adaptable, intuitive to use and more natural in the execution of activities of daily living with respect to rigid hands \cite{stephens2019survey,CapsiMorales}, they do reduce vibration transmission, demoting perception. Therefore, for users utilizing advanced prosthetic designs or those who still desire tactile feedback, it's crucial to recognize that feedback device designs should consider the natural vibration transmission through the prosthesis.
Although less intense than in rigid designs, this transmission remains persistent and offers potential for enhancement, enriching user experience without adding artificial signals unrelated to natural vibration transmission.
One potential approach may involve enhancing natural vibration transmission by integrating vibrotactile actuators within the socket controlled by accelerometer data and adjustable gain settings. Furthermore, we speculate that training a neural network to identify impacts based on socket accelerations from various prosthetic hands could lead to the development of a device that is independent of the specific type of prosthesis used, offering a more adaptable and resilient solution. Nevertheless, we acknowledge the difficulty of deciding which hand propagation pattern to prioritize. Overall, these potential solutions serve the same purpose of utilizing existing vibrations and potentially decreasing users' cognitive load required to interpret artificial tactile information transmitted by tactile feedback devices. Further research is necessary to fully comprehend the phenomenon and effectively utilize the results obtained, and we intend to address these challenges in future investigations.
Additionally, results show that detecting and recognizing tactile cues from the accelerometers at the bottom of the socket is feasible. This indicates that valuable data can be extracted from a region between the prosthetic hand and the end of the socket. Considering that haptic devices for prosthetics currently on the market have sensors placed on the prosthetic hand fingers, we believe that placing them on the socket could enhance the robustness of the prosthesis, reducing the risk of breakage or malfunctions, and increase the wearability and integration of this type of devices.
\section{Conclusions} \label{sec8}
This work successfully tested tactile perception and transmission in four prosthetic hands with different characteristics. 
The results proved that a prosthetic user perceived tactile stimuli with the four bionic hands without a haptic system.
 Interestingly, finger discrimination performances were all higher than the chance level, with decreased performances for the advanced hands.
The four bionic hands were characterized by impact forces and high-frequency stimuli transmission through the socket upon impacts with each finger.
 It has been quantitatively shown that rigid hands promote tactile transmission while more advanced hands absorb impacts.
We have shown that ANNs trained on signals from vibration sensors on the socket can recover maximum finger discrimination ability even with softer articulated hands. 
This indicates that rigid prosthetic designs favour natural vibration transmission, potentially reducing the need for extra tactile feedback devices, while advanced designs may lack this phenomenon, presenting an opportunity to enhance weaker tactile transmission and create intuitive haptic devices that minimize adaptation to artificial signals.
    \vspace{-0.3cm}
\section*{ACKNOWLEDGMENT}
The authors would like to thank Manuel Barbarossa,  Mattia Poggiani, Marina Gnocco and Emanuele Sessa for their valuable support in the experiments. 
    \vspace{-0.4cm}
\bibliographystyle{IEEEtran}
\bibliography{mybibfile}
    \vspace{-0.5cm}
\section{Biography}
    \vspace{-1.2cm}
\begin{biographynophoto}
{Alessia S. Ivani} received the bachelor's degree in Biomedical Engineering from Politecnico di Torino, Turin, Italy, in 2018, and the master’s degree in Biomedical Engineering from Politecnico di Milano, Milan, Italy, in 2020.
She is currently working toward the Ph.D. degree in robotics with the University of Pisa (Pisa, Italy) at the Soft Robotics Lab for Human Cooperation and Rehabilitation, IIT, Genoa, Italy.
Her research interests include prosthetics, haptics, and human-robot interaction.
\end{biographynophoto}
    \vspace{-1.2cm}
\begin{biographynophoto}
{Manuel G. Catalano} received the master’s degree in mechanical engineering and the doctoral degree in robotics from the University of Pisa, Pisa, Italy.
He is currently a Researcher with the Italian Institute of Technology, Genoa, Italy, and a Collaborator with the Research Center “E. Piaggio,” University of Pisa. His research interests include the design of soft robotic systems, human-robot interaction and prosthetics. Dr. Catalano won the Georges Giralt Ph.D. Award, the prestigious annual European award given for the best Ph.D. thesis by euRobotics AISBL.
\end{biographynophoto}
    \vspace{-1.2cm}
\begin{biographynophoto}
{Giorgio Grioli}received his PhD in Robotics, Automation and Bio-Engineering from the University of Pisa in 2011, with a thesis on identification for control of variable impedance actuators. He is the author of more than 130 scientific papers published in scientific journals and international conference proceedings in the fields of soft robotic actuation, robotic hand design, haptics, and human-machine interaction. Co-inventor of 5 robotic devices, helped found a spin-off company. He has served as Associate Editor for the ICRA and ICORR conferences (since 2015) and as Editor for MDPI – Actuators, Cambridge – Robotics, and Springer IJRR journals. Over the years, he supervised the development of 40 master's theses in the Automation Engineering and Mechanical Engineering courses and several bachelor's theses and student projects for the Robotics course. Member of the information engineering doctoral board of the University of Pisa, where he supervised 6 students and is supervising another 5. He also supervised a PhD student in Smart Industries and is supervising two PhD students of national interest in Robotics and Intelligent Machines. Since September 2023, he is a Senior Researcher at the University of Pisa, where he co-teaches "Robot Control" for the "Robotic and Automation Engineering" master's degree course and "Automatic Control" for the "Vehicle Engineering" master's degree course.”
\end{biographynophoto}
    \vspace{-1.2cm}
\begin{biographynophoto}
{Matteo Bianchi} received the B.Sc. and M.Sc. degrees in biomedical engineering and the Ph.D. degree in automatics, robotics, and bioengineering from the University of Pisa, Pisa, Italy, in 2004, 2007, and 2012, respectively. He is currently an associate professor with the University of Pisa, Department of Information Engineering, Interdepartmental Centre for Bioengineering and Robotics Research E. Piaggio. His research interests include haptic interface design, with applications in medical robotics and assistive-affective human-robot interaction, human and robotic hands: optimal sensing and control, human-inspired control for soft robots, psychophysics and mathematical modeling of the sense of touch and human manipulation.
\end{biographynophoto}
    \vspace{-1.2cm}
\begin{biographynophoto}
{Yon Visell} received the B.A. degree in physics from Wesleyan University, Middletown, CT, USA, the M.A. degree in physics from University of Texas-Austin, Austin, TX, USA, and the Ph.D. degree in electrical and computer engineering from McGill University, Montreal, QC, Canada, in 2011. He is currently an Associate Professor of Media Arts and Technology Program with the University of California, Santa Barbara Santa Barbara, CA, USA, Department of Electrical and Computer Engineering, and Department of Mechanical Engineering (by courtesy) – where he directs the RE Touch Lab, UC Santa Barbara. From 2013 to 2015, he was an Assistant Professor of electrical and computer engineering with Drexel University, Philadelphia, PA, USA. From 2011 to 2012, he was a Post-doctoral Fellow with the Institute of Intelligent Systems and Robotics, Sorbonne University, Paris, France. He was in industrial R\&D for sonar, speech recognition, and music DSP with several technology companies. His research interests include haptics, robotics, and soft electronics.
\end{biographynophoto}
    \vspace{-1.2cm}
\begin{biographynophoto}
{Antonio Bicchi} graduated from the University of Bologna, Bologna, Italy, in 1988.
He was a Postdoctoral Scholar with M.I.T. Artificial Intelligence lab, in 1988/1990. 
He is currently a Professor of robotics with the University of Pisa, Pisa, Italy, and Senior Scientist with the Italian Institute of Technology in Genoa, Italy. He teaches Robotics and Control Systems in the Department of Information Engineering (DII), University of Pisa. He leads the Robotics Group at the Research Center “E. Piaggio,” University of Pisa since 1990, where he was Director from 2003 to 2012. He is also the Head of the Soft Robotics Lab for Human Cooperation and Rehabilitation, IIT in Genoa. His research interests include robotics, haptics, and control systems.
\end{biographynophoto}
\vfill
\end{document}